\documentclass[10pt,journal,compsoc]{IEEEtran}
\usepackage{amsmath,amsfonts}
\usepackage{algorithmic}
\usepackage{algorithm}
\usepackage{array}
\usepackage[caption=false,font=normalsize,labelfont=sf,textfont=sf]{subfig}
\usepackage{textcomp}
\usepackage{stfloats}
\usepackage{url}
\usepackage{verbatim}
\usepackage{cite}
\usepackage{multirow}
\usepackage{booktabs}
\usepackage{graphicx}
\usepackage{color}
\usepackage{makecell}
\usepackage{tablefootnote}

\begin{document}

\title{Foundation of Intelligence: Review of Math Word Problems from Human Cognition Perspective
}

\author{Zhenya~Huang,~\IEEEmembership{Member,IEEE}, Jiayu~Liu, Xin~Lin, Zhiyuan~Ma, Shangzi Xue, Tong Xiao, Qi~Liu,~\IEEEmembership{Member,IEEE}, 
Yee Whye Teh,
Enhong~Chen,~\IEEEmembership{Fellow,IEEE}
\IEEEcompsocitemizethanks{\IEEEcompsocthanksitem Z. Huang,  X. Lin, Z. Ma, S. Xue, T. Xiao, E. Chen (Corresponding Author) are with the Anhui Province Key Laboratory of Big Data Analysis and Application, School of Computer Science and Technology, University of Science and Technology of China, and with the State Key Laboratory of Cognitive Intelligence, Hefei, Anhui 230000, China. Email: \{huangzhy,cheneh\}@ustc.edu.cn, \{linx,zhyma,xueshangzi,tongxiao2002\}@mail.ustc.edu.cn.
\IEEEcompsocthanksitem J. Liu, Q. Liu are with the School of Artificial Intelligence and Data Science, University of Science and Technology of China, and with the State Key Laboratory of Cognitive Intelligence, Hefei, Anhui 230000, China. Email: jy251198@mail.ustc.edu.cn, qiliuql@ustc.edu.cn.
\IEEEcompsocthanksitem YWT is with the Department of Statistics, University of Oxford, Oxford OX1 3LB, United Kingdom. Email: y.w.teh@stats.ox.ac.uk.}}

\markboth{Journal of \LaTeX\ Class Files,~Vol.~14, No.~8, August~2021}%
{Shell \MakeLowercase{\textit{et al.}}: A Sample Article Using IEEEtran.cls for IEEE Journals}

\IEEEpubid{0000--0000/00\$00.00~\copyright~2021 IEEE}

\maketitle

\begin{abstract}
\color{black} Math word problem (MWP) serves as a fundamental research topic in artificial intelligence (AI) dating back to 1960s. This research aims to advance the reasoning abilities of AI by mirroring the human-like cognitive intelligence. The mainstream technological paradigm has evolved from the early rule-based methods, to deep learning models, and is rapidly advancing towards large language models. However, the field still lacks a systematic taxonomy for the MWP survey along with a discussion of current development trends. Therefore, in this paper, we aim to comprehensively review related research in MWP solving through the lens of human cognition, to demonstrate how recent AI models are advancing in simulating human cognitive abilities. Specifically, we summarize 5 crucial cognitive abilities for MWP solving, including \emph{Problem Understanding}, \emph{Logical Organization}, \emph{Associative Memory}, \emph{Critical Thinking}, and \emph{Knowledge Learning}. Focused on these abilities, we review two mainstream MWP models in recent 10 years: neural network solvers, and LLM based solvers, and discuss the core human-like abilities they demonstrated in their intricate problem-solving process. Moreover, we rerun all the representative MWP solvers and supplement their performance on 5 mainstream benchmarks for a unified comparison. To the best of our knowledge, this survey first comprehensively analyzes the influential MWP research of the past decade from the perspective of human reasoning cognition and provides an integrative overall comparison across existing approaches. We hope it can inspire further research in AI reasoning. Our repository is released on \url{https://github.com/Ljyustc/FoI-MWP}.

\end{abstract}


\section{Introduction}
\IEEEPARstart{M}{athematical} {\color{black} {reasoning is a fundamental aspect of human cognition and serves as a critical benchmark to assess the level of artificial intelligence (AI) systems~\cite{davies2021advancing}. Among various mathematical tasks, Math Word Problems (MWPs) stand out as a foundational branch, where researchers show a consistent focus dating back to the 1960s~\cite{zhang2020gap}. As shown in Table~\ref{tab:mwp_example}, solving an MWP typically requires models to understand a problem text in natural language and then reason a formal mathematical expression/rational to derive the answer. This procedure mirrors several basic human cognitive functions, making MWPs a valuable tool to study and advance the cognitive reasoning abilities of AI.}}

\begin{table}[!t]
    \caption{An Example of a Math Word Problem\label{tab:mwp_example}}
    \centering
    {\renewcommand{\arraystretch}{2} 
    \begin{tabular}{|c|c|}
         \hline
         \textbf{Problem} & \makecell[l]{Weng earns \$12 an hour for babysitting. Yesterday, \\ she just did 50 minutes of babysitting. How much \\ did she earn?} \\
         \hline
         \textbf{Answer} & 10 \\
         \hline
         \textbf{Expression} & \makecell{$12\ /\ 60\ \times\ 50$ (infix)\\ $12\ 60\ /\ 50\ \times$ (postfix) \\ $\times\ /\ 12\ 60\ 50$ (prefix)}\\
         \hline
         \makecell{\textbf{Expression} \\ \textbf{Tree}} & \makecell{\vspace{-8pt}\\\includegraphics[width=0.2\linewidth]{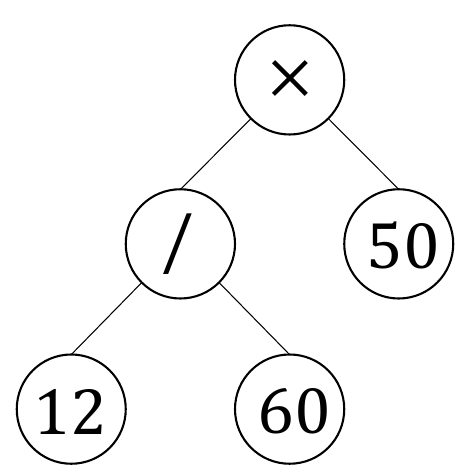}} \\
         \hline
         \textbf{Rationale} & \makecell[l]{Weng earns $12 / 60 = 0.2$ per minute. Working $50$ \\ minutes, she earned $0.2 \times50 = 10$.} \\
         \hline
    \end{tabular}
    }
\end{table}
{\color{black} {To overcome this task, researchers have proposed many representative approaches, where the technical paradigm has roughly evolved with three types. Specifically,}} earlier methods for solving MWPs have largely relied on manually crafted rules and templates to parse and solve problems~\cite{bakman2007robust}. While effective to a certain extent, these methods are limited by their dependence on predefined patterns and lack the flexibility to handle the diverse and nuanced nature of real-world problems. In recent years, deep learning has emerged as the dominant methodology for tackling MWPs~\cite{wang2017deep,wang2019template,shi2015automatically,huang2018neural,shen2021generate}. This shift has led to various innovative approaches, each focusing on different aspects of the problem-solving process. The first significant portion of the research concentrates on better modeling the information presented in problem text~\cite{zhang2020graph,li2019modeling,qin2021neural} (e.g., syntactic structures~\cite{lin2021hms}), integrating external knowledge (e.g., commonsense knowledge~\cite{wu2020knowledge}, mathematical formulas~\cite{yang2022logicsolver,liu2023guiding}), or employing advanced pre-trained language models (e.g., BERT~\cite{koncel2016mawps}, GPT~\cite{brown2020language}), to enhance the model's comprehension abilities. The second line of research develops heuristic reasoning patterns such as tree-structured reasoning~\cite{xie2019goal,wang2018translating,wang2022structure} or directed acyclic graph (DAG) reasoning~\cite{cao2021bottom,koncel2016mawps} to enhance the interpretability and accuracy of the reasoning process. As large language models (LLMs) like GPT-4 achieving impressive results in many machine learning tasks~\cite{wei2022chain,zhao2023survey,wang2023document,zhang2024vision}, there is also a growing interest in applying LLMs to MWPs~\cite{zhu-etal-2023-solving,liang-etal-2023-gpt}. 

{\color{black} {Despite these advancements, some pioneers have dedicated the survey on promoting MWP research~\cite{zhang2020gap,lu2023survey}. However, existing literature exhibits several technical and conceptual limitations. Technically, existing literature are either outdated~\cite{zhang2020gap} (to the best of our knowledge, the most recent MWP survey has been published in 2020~\cite{zhang2020gap}), focusing primarily on reasoning in formal languages and including only early works, or they are overly broad, investigating all mathematical reasoning tasks without delving into the detailed development of MWPs~\cite{lu2023survey}. Conceptually, these surveys generally summarize the research advancements from the technical perspective of model structures (e.g., Seq2Seq-based, Graph-based), which do not examine their human-like intelligence capabilities required for MWP-solving. As a long-standing and fundamental research topic, it is of even great significance to explore the level of cognitive reasoning capabilities that the AI models can achieve on MWPs. Especially for LLMs, some studies suggest that their cognitive abilities are close to that of an 8-year-old child~\cite{zador2023catalyzing}. However, how these models simulate human intelligence remains great value but still exists an open discussion. In particular, when it comes to solving complex MWP problems, replicating the way humans think is still a challenge that requires further exploration. In summary, there is a lack of systematic taxonomy of MWP survey, which focus on discussing the development trend of their technical reasoning capabilities.

Our review aims to fill this gap by examining MWP-related research through the lens of human cognition. We provide a comprehensive overview of how current artificial intelligence models are advancing in simulating human cognitive abilities, and to reveal the intersection of these methods with human intelligence. This perspective not only enriches the field's  understanding of MWPs but also offers insights into developing more sophisticated AI systems capable of genuine mathematical reasoning.}}

Specifically, based on cognitive science theories, {\color{black} {in Figure.~\ref{fig:over_nn},}} we elaborate on fives crucial cognitive abilities that current methods mainly enhance: \emph{Problem Understanding}, \emph{Logical Organization}, \emph{Associative Memory}, \emph{Critical Thinking}, and \emph{Knowledge Learning}. Our analysis reveals that current efforts primarily focus on the first two foundational abilities-\emph{Problem Understanding} and \emph{Logical Organization}, with fewer studies addressing higher-order cognitive abilities like \emph{Knowledge Learning} and \emph{Critical Thinking}. Additionally, since LLMs unify various human-like abilities, we especially conduct a detailed review of current LLM-based work. We find that efforts to enhance the five aforementioned cognitive abilities for LLMs are not evenly distributed, revealing a feasible direction for future optimization by incorporating more cognitive capabilities. Last but not least, considering that existing work (whether small or large models) has not been widely tested on MWP datasets, we supplement this study by comprehensively evaluating representative small models, LLMs (e.g., GPT-3.5 and LLaMA3.1-8B), and LLM-based methods (e.g., CoT~\cite{wei2022chain}, ToT~\cite{yao2024tree}, GoT~\cite{besta2024graph}, PoT~\cite{chen2023program}, PAL~\cite{gao2023pal}) on five widely-used MWP datasets.

To sum up, researchers in the
community can benefit from this survey in the following ways:
\begin{itemize}
    \item From a novel perspective of cognitive abilities, we systematically and comprehensively review existing research on MWPs. This provides crucial support for positioning the cognitive level of current AI models.
    \item In addition to reviewing small models, we also analyze the cognitive abilities of current LLMs on MWP task, offering insights for further developing the capabilities of these large models.
    \item In the experimental section, we supplement the performance of {\color{black} {various mainstream models}} across {\color{black}{5 benchmarks}}, including 14 small MWP models, 4 representative LLMs, and 10 state-of-the-art LLM-based methods, clearly illustrating the impact of different cognitive abilities on reasoning accuracy.
\end{itemize}

The remainder of the paper is organized as follows. We first introduce the definition of MWP task and discuss the five types of human cognitive abilities involved in Section 2. Then, we classify current small-scale models according to different cognitive abilities in Section 3. In Section 4, we summarize existing analysis of LLMs on MWP task. Section 5 introduces representative MWP datasets and compares existing models' performances. In Section 6, we further expand on representative mathematical reasoning tasks involving other cognitive abilities. We conclude the paper and point out several future directions in MWPs that merit further examination in the final section.

\section{Task Definition}
Math word problems (MWPs) represent a fundamental task in mathematical reasoning, designed to solve primary school–level problems that require basic reasoning
As shown in Table~\ref{tab:mwp_example}, the input is a problem $P$ composed of natural language (e.g., ``an hour'') and quantities (e.g., ``12'').
The output is a numeric answer $A_P$ (e.g., ``10'').

To derive the answer, humans typically represent the reasoning process in two forms: a mathematical expression or a rationale. The mathematical expression $E_P$ encodes the reasoning in a formal symbolic form (``$12/60\times50$''), which can be defined as a sequence of symbols of mathematical operators $V_O$ (e.g., $\{+,-,\times,/\}$), constants $V_C$ ($\{1, \pi, 60\}$), and quantities from the problem $N_P$ ($\{12, 50\}$), i.e., $E_P=\{y_1, y_2, ..., y_m\} (y_i\in V_O \cup V_C \cup N_P)$. In contrast, the rationale $R_P$ combines symbolic expressions with natural language explanations to provide a precise and comprehensible reasoning process, e.g., ``Weng earns $12/60=0.2$ per minute ...''. Generally, the rationale has higher interpretability, but requires larger vocabulary and more powerful models to generate. Given a problem $P$, the MWP task aims to first generate either the mathematical expression $E_P$ or the rationale $R_P$, and then compute the final answer $A_P$.

Recalling how humans perform the above reasoning process, we find that it requires several key cognitive abilities~\cite{mayer2012process,phonapichat2014analysis,daroczy2015word,pongsakdi2020makes}. For example, to answer the problem in Table~\ref{tab:mwp_example}, we first need problem understanding to extract the goal and given information from the problem. Next, logical organization is required to structure the reasoning steps, such as first computing ``how much Weng earns per minute'' and then ``how much she earns in total''. During this process, knowledge learning is needed to apply relevant facts (e.g., ``an hour equals 60 minutes''). For more complex problems, additional abilities become essential, including associative memory to recall prior experiences and critical thinking to evaluate alternative reasoning strategies. Building on findings from psychology and cognitive science, we summarize five key cognitive abilities required to solve MWPs: \emph{problem understanding}~\cite{mayer2012process,phonapichat2014analysis}, \emph{logical organization}~\cite{xie2019goal,daroczy2015word,pongsakdi2020makes}, \emph{associative memory}~\cite{anderson2014human,raaijmakers1981search}, \emph{critical thinking}~\cite{johnson2002neural,grant2002self,astington201316}, and \emph{knowledge learning}~\cite{liu2023guiding,hospedales2021meta,guo2021context}.

When bridging human and machine intelligence, existing MWP solvers typically emulate certain cognitive abilities to achieve accurate solutions. These solvers can be broadly divided into neural network (NN)-based solvers and large language model (LLM)-based solvers. NN-based solvers, constrained by limited parameters and annotated data, usually focus on a single cognitive ability and generate mathematical expressions as outputs~\cite{wang2017deep,xie2019goal,lin2021hms,liu2022cognitive}. LLM-based solvers leverage massive pre-training corpora and abundant parameters to capture richer abilities, often producing natural language rationales for MWPs~\cite{wei2022chain,kojima2022large,zhouleast}. In the following sections, we first review related works in traditional NN-based solvers, and then introduce recent advances in LLM-based solvers. The taxonomy of the methods introduced in this paper is presented in Figure~\ref{fig:over_nn}.

\section{Neural Network Solvers}
\label{sec:neural_solvers}
\begin{figure*}[ht]
\setlength{\abovecaptionskip}{1pt}
\centering
\includegraphics[width=0.95\textwidth]{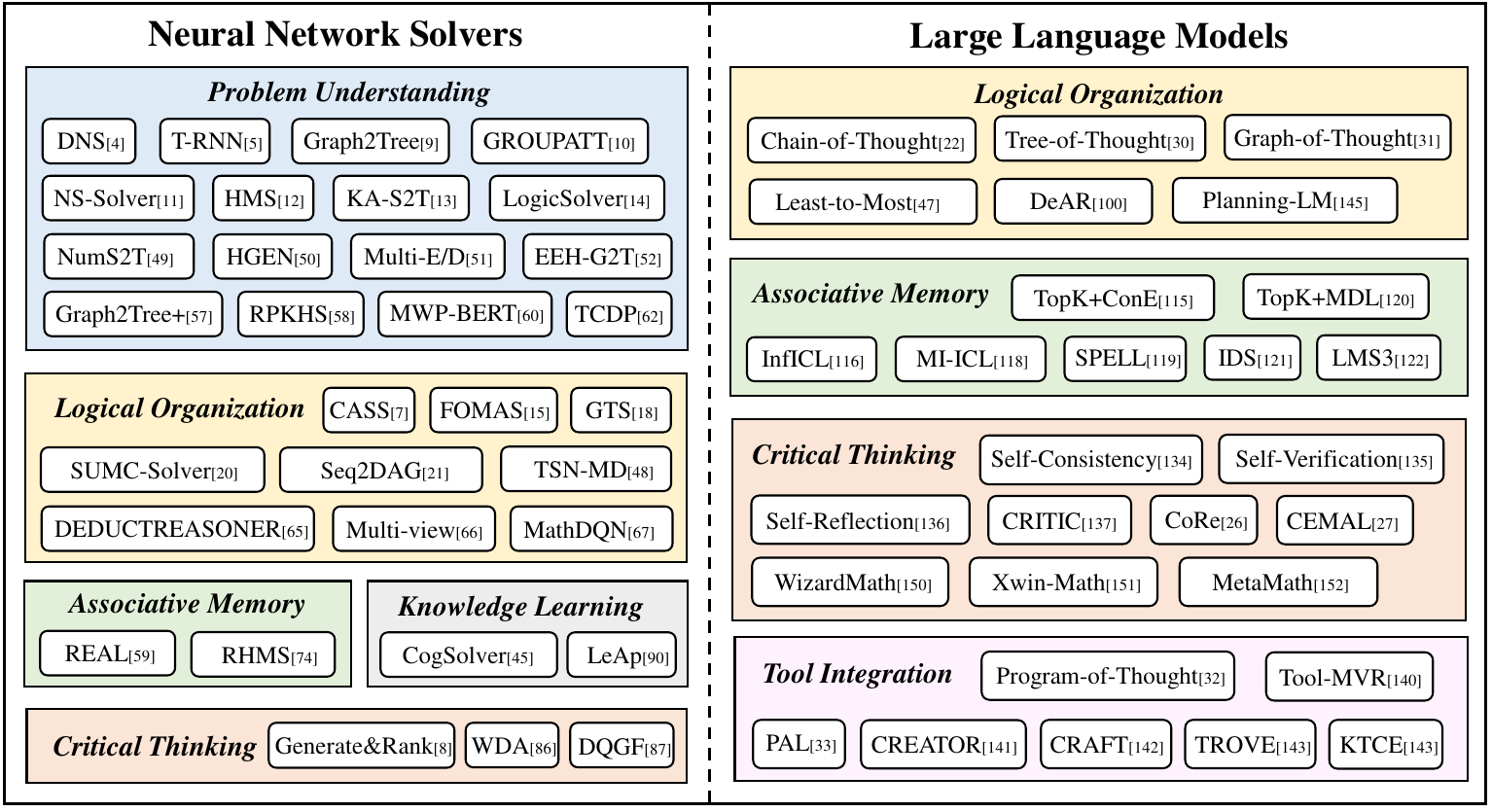}
\centering
\caption{Taxonomy of Methods Based on Human Cognitive Abilities.}
\label{fig:over_nn}
\vspace{-10pt}
\end{figure*}

In this section, we review NN-based solvers, which imitate specific cognitive abilities and generate mathematical expressions as solutions. We organize the taxonomy according to the five cognitive abilities in Figure~\ref{fig:over_nn}, including problem understanding, logical organization, associative memory, critical thinking, and knowledge learning.

\subsection{Ability 1: Problem Understanding}

Accurately understanding the problem is the basis of MWP solving~\cite{mayer2012process,phonapichat2014analysis}. To correctly answer the MWP, the model needs to first extract the inherent information in the problem, such as the semantics and the quantitative relationships. In addition, external knowledge beyond the problem statement, such as commonsense or domain-specific knowledge, is also necessary in problem understanding. To this end, researchers have proposed several methods to capture both inherent and external information to enhance problem comprehension.

\noindent\textbf{Inherent Information.} The semantics is the most important information in the problem. In early studies, researchers model the problem as a sequence of words, and capture its semantics with RNN-based encoders. To the best of our knowledge, DNS~\cite{wang2017deep} is the first to apply deep learning technique in MWP solving following the encoder-decoder framework. DNS extracts the problem semantics into a hidden vector with a GRU encoder upon word embedding, which is later used to generate the expression in decoder. The sequence-based encoder is widely used in various MWP solvers and further improved~\cite{wang2019template, xie2019goal, zhang2021teacher, wang2018translating}. For example, Li et al.~\cite{li2019modeling} enhanced the encoder with group attention mechanism to further incorporate different relationships in the problem.
In recent studies, researchers have started diving into the sophisticated language structure of the problem, and modeling the problem semantics following human-like reading. Lin et al.~\cite{lin2021hms} proposed the HMS framework for the hierarchical language structure of the problem, i.e., the problem is composed of several clauses, and each clause is further composed of words forming the dependency tree structure. To this end, HMS designs the word-clause-problem hierarchical encoder to capture fine-grained semantics in different levels, and further captures the dependency relationship between words in each clause for precise semantics understanding. 

Quantitative information is another key component in MWP. In the literature, there are two lines of methods to capture the quantity features, (1) modeling the relationships between quantities and words, and (2) learning the literal values of quantities. For the former, to capture the quantity relationships, Zhang et al.~\cite{zhang2020graph} constructed the quantity cell graph and quantity comparison graph. The quantity cell graph adds undirected edges between quantity and related descriptive words to provide necessary semantics (e.g., ``12'' is associated with ``earns'', ``\$'', ``hour'' in Table~\ref{tab:mwp_example}), and the quantity comparison graph adds directed edges between quantities from the larger one to the smaller one to capture the numerical relationships (``50'' is larger than ``12''), as it is more likely to subtract a smaller quantity from a larger one. The two graphs are embedded with a graph transformer to enrich problem understanding. For the latter, to learn the quantity values, NumS2T~\cite{wu2021math} applies a digit-to-digit encoder on the literal contents to obtain quantity representations, since the symbols and values of quantities also affect problem understanding (e.g., ``25 more hours'' and ``25\% more hours'' are totally different). NumS2T also designs three losses for quantity encoding by comparing quantity value, predicting quantity type and whether it is used in expression. Similarly, Qin et al.~\cite{qin2021neural} proposed to predict the number and location of quantities, and required constants from commonsense (e.g., ``an hour equals to 60 minutes'') to encourage quantity learning.

Researchers have also tried to ensemble various inherent information and modeling methods to achieve better problem understanding. For example, Zhang et al.~\cite{zhang2022hgen} designed a heterogeneous graph composed of dependency parsing tree, quantity cell graph and quantity comparison graph, and employed a type-aware graph embedding networks to learn the problem representations. Shen et al.~\cite{shen2020solving} proposed a Multi-E/D framework which first exploited an sequence-based encoder to obtain the context representation, and then employed two GCNs on dependency parsing tree and quantity comparison graph to learn the semantic and quantity features for enhanced problem understanding.

\noindent\textbf{Explicit Knowledge.} Apart from the inherent information from the problem, explicit knowledge stored in external knowledge graphs (KGs) and textual records is also necessary to provide absent common information, such as commonsense and domain-specific knowledge. 

To leverage KGs in MWP solvers, a representative approach is KA-S2T~\cite{wu2020knowledge}, which exploits word–category commonsense graphs. Specifically, KA-S2T constructs an entity graph where nodes represent words in the problem and their shared categories (e.g., both “red” and “green” belong to the category “color”), with edges linking words to their corresponding categories. To further integrate phrase-level knowledge, KA-S2T introduces phrase category nodes for phrases that share the same category and word (e.g., both “red apple” and “green apple” correspond to “color+apple”), and connects these nodes to the first and last words of the phrase. Based on the constructed entity graph, KA-S2T applies a Graph Attention Network (GAT) to learn node and problem representations. Wu et al.~\cite{wu2021edge} improved KA-S2T by incorporating richer relationships among words, and proposed EEH-G2T that integrates self–neighbor relations, same-word relation, dependency relations, and external category relations into a heterogeneous graph to enhance understanding.

Another widely used form of knowledge is natural text records, such as domain-specific mathematical formulas~\cite{roy2018mapping,liu2023guiding,yang2022logicsolver} (e.g., ``$cost=price\times quantity$''). Yang et al.~\cite{yang2022logicsolver} proposed LogicSolver to incorporate textual formula knowledge for interpretable MWP solver. They first constructed an InterMWP dataset by annotating the formula required in each step. Based on the dataset, LogicSolver trains a BERT-based retriever to locate highly-correlated formulas from a formula set. Given one problem, LogicSolver first retrieves related formulas with the retriever and then encodes both the problem and the retrieved formulas with a BERT encoder. By incorporating the formula in this way, the model improves not only problem-solving accuracy but also interpretability.

\noindent\textbf{Parametric Knowledge.} In addition to explicit knowledge stored in KGs and texts, world knowledge can also be stored in the parameters of pre-trained language models (PLMs) such as BERT and RoBERTa~\cite{devlin2019bert,Liu2019RoBERTaAR,petroni2019language}.
Therefore, PLMs are widely used in MWP solvers to leverage their parametric knowledge for problem understanding. A straightforward way is to incorporate them directly in the encoder or decoder modules. For example, Kim et al.~\cite{kim2022improving} replaced the traditional word2vec embeddings with pre-trained RoBERTa. Yu et al.~\cite{yu2021improving} and Huang et al.~\cite{huang2021recall} adopted BERT and RoBERTa in the problem encoder to exploit implicit commonsense knowledge.

However, existing PLMs are pre-trained on general corpus, and thus may not fully capture MWP-specific features (e.g., limited sensitivity to quantities). To this end, researchers have proposed several MWP-specific training tasks and methods to further promote their capabilities in MWP understanding. For example, Shen et al.~\cite{shen2021generate} fine-tuned a pre-trained BART as the MWP solver. To improve the PLMs on capturing the vital numeracy information, Liang et al.~\cite{liang2022mwp} pre-trained an MWP-BERT on MWP dataset with quantity-related tasks, such as quantity counting, quantity type prediction, answer type prediction, as well as the original masked language modeling. 
Li et al.~\cite{li2022seeking} proposed to perceive the divergence of MWP patterns via contrastive learning. Given a problem $p$, positive examples are problems whose template expressions (i.e., expressions with operators and placeholder quantities) are identical to $p$ or contain sub-structures matching $p$, which means that they share the same quantity relationships. Hard negative examples for $p$ are problems with the same number of nodes in their template expressions but different operators. In this way, the PLMs could be aware of the underlying quantity patterns of the problems. Similarly, Qin et al.~\cite{qin2023template} employed an auxiliary loss to distill knowledge from a teacher encoder to a student encoder. This loss pulls closer problem representations with the same template expression while pushing apart those with different templates, both in the student feature space and the student–teacher cross-feature space.

\subsection{Ability 2: Logical Organization} 
Humans often organize mathematical reasoning using different logical structures, commonly in the form of sequences, trees, or DAGs.
{\color{black}{Such structures help arrange mathematical elements according to their inherent relationships, facilitating a clearer and more systematic reasoning process. Motivated by this, }}researchers have designed various methods to generate expressions that follow specific logical forms, which is usually implemented by the decoder within the classical encoder-decoder framework. 

\noindent\textbf{Sequence-based.} The most straightforward way to organize the expression is a sequence of symbols (quantities and operators), where each symbol is predicted autoregressively based on the problem and the previously generated symbols. As shown in Table~\ref{tab:mwp_example}, the expression can be represented in prefix, postfix or infix form, depending on the relative positions of the operators and its operands in each step. Among these, the postfix form is most widely used due to its ease of handling. Early studies~\cite{wang2017deep, li2019modeling, wang2018translating} employed classical LSTMs with attention mechanism as the decoder to generate the symbol sequence of expressions. They also utilized auxiliary rules to filter out illegal symbols in each position to ensure valid expressions~\cite{wang2017deep}. Wang et al.~\cite{wang2019template} further enhanced the vanilla decoder with a two-stage method, where the decoder first generates an expression template with placeholder operators, and then predicts each operator based on the representations of the two operands using a recursive network.

\noindent\textbf{Tree-based.} Sequence-based methods often struggle to handle the complex structure of expressions and can not guarantee the legality of expressions. To address this, a more reasonable method is to organize the expressions as expression trees (Table~\ref{tab:mwp_example}), which inherently ensure validity. In such trees, the leaf nodes are the quantities, and the inner nodes are the operators. Each expression has a one-to-one correspondence with an expression tree, which explicitly reflects the computation process and effectively captures structural features. Consequently, most advanced neural MWP solvers convert the expression generation into the tree construction (i.e., the Seq2Tree paradigm), which is often implemented following the depth-first top-down recursive manner from the root node to the leaves. If the predicted symbol at a node is an operator, the solver moves to its left child and continues. Otherwise, it backtracks to the nearest right child  until the tree is complete. In this way, the predicted symbols will naturally form a legal prefix expression as the output.

Liu et al.~\cite{liu-etal-2019-tree} conducted one of the earliest studies on tree-based decoders by incorporating parent and sibling information. Motivated by the goal-driven problem solving of humans, Xie et al.~\cite{xie2019goal} proposed the most famous tree-based decoder GTS. The key insight is that each node in the expression tree corresponds to a goal (e.g., ``How much did she earn in total?'' for the root in Table~\ref{tab:mwp_example}). A goal can either be directly fulfilled by a single quantity or decomposed into two sub-goals (e.g., ``How much did she earn per minute?'' and ``How many minutes did she work?''), which are connected by one operator (``$\times$''). To implement this idea, GTS assigns a goal vector $q$ to each node and predicts the symbol $y$ based on $q$. For the root node, the goal $q$ is initialized with the hidden vector from the encoder. For other nodes, $q$ is recursively derived from the parent goal, taking into account the parent operator and, when applicable, the left sibling subtree. 
The goal-driven decoder has since become a foundation for many MWP solvers~\cite{zhang2020graph, zhang2021teacher, yang2022logicsolver, lin2021hms}, with researchers proposing various extensions. For example, Zhang et al.~\cite{zhang2021teacher} distilled knowledge from a teacher network to a multi-decoder student network based on GTS to encourage diversity. Wu et al.~\cite{wu2020knowledge, wu2021edge} introduced a state aggregation mechanism to incorporate partially generated expression trees. Hong et al.~\cite{hong2021learning} further proposed a weakly-supervised learning-by-fixing training framework under answer supervision to correct erroneous solutions in the expression trees.

To further guide goal decomposition and symbol prediction with auxiliary information, Liu et al.~\cite{liu2023guiding} designed the FOMAS framework that incorporates the formula knowledge. 
In symbol prediction, given a goal vector $q$, FOMAS applies three mechanisms to guide prediction with formulas: (1) a formula-selected mechanism, which selects a new formula based on goal vector and formula embeddings, and takes the top operator as the predicted symbol; (2) a formula-inherited mechanism, which inherits corresponding operator or concept of the formula selected by the parent to predict the symbol; and (3) a direct reasoning mechanism, which predicts the symbol without formula information. FOMAS applies a confidence-based ensemble on the three mechanism to determine the final result. In goal decomposition, FOMAS generates child goal vectors based on the corresponding concept in the formula, which provides better interpretability for each reasoning step.
\begin{figure}[!t]
\setlength{\abovecaptionskip}{1pt}
\centering
\includegraphics[width=0.97\linewidth]{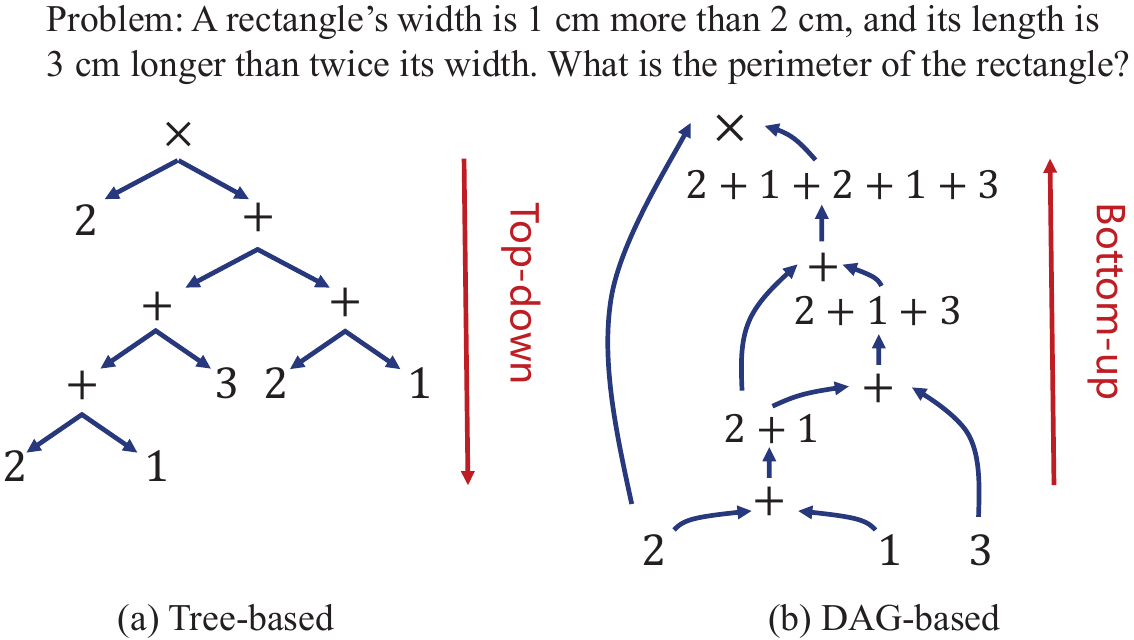}
\caption{Comparison between Tree-based and DAG-based methods.}
\label{Compare_tree_dag}
\end{figure}

Classical binary expression trees are order-dependent, so that they can not effectively handle the equivalent expressions with simple commutation and deformation. For example, ``$(12/60)\times50$'' and ``$50\times(12/60)$'' are equivalent, but they must be expressed differently in a binary tree.
To overcome this limitation, SUMC-Solver~\cite{wang2022structure} designs an order-independent tree, termed the M-tree, which allows arbitrary branching to more flexibly represent expressions. SUMC-Solver reformulates the ordinary binary operators $+$, $-$, $\times$, and $/$ into order-independent multivariate operators: accumulation and multiplication, augmented with negation and reciprocal operations. Based on these operators, SUMC-Solver adopts the M-tree with arbitrary branches to organize the expression, which is rooted on accumulation. By assigning lower priority to accumulation and treating sibling nodes as order-independent, M-tree provides a normalized way to represent families of equivalent expressions, thereby reducing the search space.

\noindent\textbf{DAG-based.} Unlike Seq2Tree, which simulates human top-down decomposition process by recursively expanding an expression tree, Seq2DAG emulates human inductive reasoning, generating a directed acyclic graph (DAG) in a bottom-up fashion. The core idea is that in each step, an operator and two operands are selected to form a sub-expression. For example, in Figure~\ref{Compare_tree_dag}(b), the first step selects operator ``$+$'' and operands ``$2$'' and ``$1$'' to form the sub-expression ``$2+1$''. This sub-expression is then added to the set of operands, allowing it to be reused in subsequent reasoning steps. Then, in the second step, the operator $+$, along with the operands ``$2+1$'' and ``$3$'', are chosen to generate the sub-expression ``$2+1 + 3$'', thereby forming a graph structure. Compared to Seq2Tree in Figure~\ref{Compare_tree_dag}(a), the key advantage of Seq2DAG is its ability to continuously store and reuse previously generated sub-expressions. In contrast, Seq2Tree might redundantly generate the same subtree multiple times (e.g., there are two identical subtrees ``$2 + 1$''). Thus, Seq2DAG not only improves reasoning rationality but also better aligns with human reasoning process. However, Seq2DAG pattern needs to consider all possible ($operator$, $operand_1$, $operand_2$) pairs at each step. As the operand set grows linearly, the candidate space expands exponentially. Therefore, the efficiency is a major bottleneck that hinders their widespread application.

In practice, DAG-based methods typically consist of two modules. The first module evaluates the probability of each candidate (operator $op$, operand $i$, operand $j$) at each step. It begins with fusing the representations of $op$, $i$, and $j$, denoted as $e_{op}, e_i, e_j$, into a joint representation $e_{i,j,op}$ for the sub-expression ($i, op, j$). A softmax classifier is then employed to select the sub-expression with the highest probability, or alternatively, a binary classifier is used to compute the selection probability for each sub-expression individually, retaining only those whose probabilities exceed a given threshold. The second module is a termination module, which determines when to stop the reasoning process. After each step of sub-expression generation, this module evaluates whether the current expression is complete and should be treated as the final result.

Cao et al.~\cite{cao2021bottom} proposed the first DAG-based method, which designs different network architectures for operators with different mathematical properties. If $op$ does not satisfy commutative law (e.g., ``$-,\div$''), they directly concatenate the representations $e_i$ and $e_j$. In contrast, for commutative operators (e.g., ``$+,\times$''), they sum $e_i$ and $e_j$ to eliminate the order dependency. The resulting representation is then linearly combined with the operator representation $e_{op}$ and fed into an LSTM cell. A binary classifier is used to independently calculate the selection probability for each ($op, i, j$) pair. The algorithm continues until none of the ($op, i, j$) pairs has a selection probability greater than 0.5.

Another prominent work is DEDUCTREASONER~\cite{jie2022learning}, which differs from~\cite{cao2021bottom} in three key aspects. First, after calculating the representation $e_{i,j,op}$, it integrates the operand information and compute a score $score_{i,j,op}$ for the sub-expression. The pair with the highest score is selected as the reasoning result for that step. Second, it introduces an additional Terminator network (Eq.~\eqref{deductive_2}) to explicitly learn the stopping condition. Third, after each reasoning step, the representation $e_{i,j,op}$ of the selected expression is used to update the representations of all other operands, which aims to prevent the repeated selection of the same ($op, i, j$) pair across different steps.
\begin{equation}\label{deductive_2}
    S = score_{i,j,op}+w_\tau \cdot FFN_s(e_{i,j,op})
\end{equation}

Beyond relying solely on a single top-down Seq2Tree or bottom-up Seq2DAG manner, Zhang et al.~\cite{zhang2022multi} further proposed a hybrid approach, which aligns the two perspectives via contrastive learning on sub-expression representations to enhance the robustness of reasoning. Shen et al.~\cite{shen2020solving} introduced a Multi-E/D approach, which leverages multiple types of decoders (as well as encoders) to improve model precision. It employs a sequence-based decoder to generate expressions in postfix order and a tree-based decoder to generate them in prefix order. The final solution is determined by selecting the expression with the highest probability output from these decoders, which can promote the model's adaptability to different problem formats.

\noindent\textbf{MDP-based.} Except from organizing logic with fixed mathematical formations like Tree or DAG, a more flexible and principled direction is to formulate expression generation as a Markov decision process (MDP) and resolve it by reinforcement learning (RL)~\cite{wang2018mathdqn,huang2018neural}. This formulation also addresses the challenge of the exponential search space of mathematical concepts~\cite{faldu2021towards}. One pioneering work is MathDQN~\cite{wang2018mathdqn}, which leverages a Deep Q-Network (DQN)~\cite{mnih2015human} to recursively construct an expression tree by selecting relevant quantity pairs and operators. The state is represented as a vector of three types of crafted features for individual quantity, pair quantities, and problems. The action is defined as selecting a pair of quantities and an operator. The reward function is straightforward: if the selected operator is correct for the quantities, a positive reward is provided, otherwise a negative reward is issued for punishment.

Instead of framing the entire solving process as RL, some works used RL as a method to optimize the solvers. Specifically, Huang et al.~\cite{huang2018neural} pointed out that the objective of answer accuracy is non-differentiable, and Maximum Likelihood Estimation (MLE) is indeed a surrogate objective aimed at maximizing equation likelihood during training. This may suffer from the issue of ``train-test discrepancy'', which can result in the generation of spurious numbers or the production of numbers in incorrect positions. To address this, they replace the training objective and gradient of MLE at each step with Eq~\eqref{cass}.
\begin{equation}\label{cass}
\begin{aligned}
    Loss_{RL} = & - \sum_{y_i\in E_P}\mathbb{E}_{p_\theta(y_i|E^{i-1}_P)}[R(y_i,E^{i-1}_P)] \\
    \nabla_\theta Loss_{RL} =& - \sum_{y_i\in E_P}\sum_{y\in Top_k}p_\theta(y|E^{i-1}_P)\cdot R(y,E^{i-1}_P)\\
    & \cdot\nabla_\theta \log p_\theta(y|E^{i-1}_P)
\end{aligned}
\end{equation}
where $R(y_i,E^{i-1}_P)=1$ if $y_i$ yields to the correct solution, and $-1$ otherwise. To avoid traversing all possible expressions, the authors approximately calculate the gradient by sampling the top-k expressions $Top_k$ in the expectation.

\subsection{Ability 3: Associative Memory}
Psychological research points out that associative memory is a critical human cognitive ability that allows individuals to recall related information from past experiences and apply it to new situations~\cite{anderson2014human,raaijmakers1981search}. For example, complementary learning system theory~\cite{mcclelland1995there,kumaran2016learning,o2014complementary} reveals how the memory replaying mechanism in human brain is implemented with the synergistic interactions between hippocampus and neocortex, which enables the transfer of knowledge across different contexts and situations.

In the context of MWP, researchers have sought to equip models with the ability to retrieve and exploit memories to support reasoning. Memory, in this case, refers to problems that the model has previously encountered, essentially forming an experience bank derived from the training set. By accessing related examples from memory, these models mimic the human process of recalling associated problems to guide the solution of new ones. One of the typical approaches is REAL~\cite{huang2021recall}, which introduces a memory-augmented solver that comprises four key modules: a memory module for retrieving similar problems based on word2vec embeddings~\cite{mikolov2013efficient}, a representation module for encoding the concatenation of the test problem and the retrieved ones, an analogy module to establish relationships between the current problem and the retrieved ones, and a reasoning module that integrates a copy mechanism to generate the solution. 

Moreover, Lin et al.~\cite{lin2023learning} introduced RHMS, which designs a meta-structure to better capture the association between problems in memory and the problem at hand. Here, we focus on the process of constructing the experience bank. In meta-structure, each problem is represented as a dependency tree where the node represent part-of-speech (POS) tag of each word, and the edge represent dependencies (e.g., subject-verb-object relations). Then, the authors extract all sub-paths from the ancestor node to the offspring in the tree as structure paths, which can emulate human's association on the logical structure of problem template. By applying TF-IDF vectorization on these paths, RHMS computes a cosine similarity between different problems, forming a structural similarity graph that links problems with similar logical forms, {\color{black}{thus effectively modeling historical similar problems as experience.}} This graph enables the solver to aggregate information from related problems using a Graph Attention Network (GAT)~\cite{velivckovic2018graph}, enhancing the semantic understanding of the target problem through its connections with problems in experience.

\subsection{Ability 4: Critical thinking}
Critical thinking is a fundamental cognitive skill extensively studied in philosophy, cognitive psychology, and education for its role in advancing human intellectual development~\cite{lai2011critical,moore2012critical,paul1992critical}. For MWP task, it involves continuously evaluating problem-solving strategies and identifying new challenges to deepen the understanding of the underlying reasoning~\cite{johnson2002neural,grant2002self,astington201316}. Consequently, fostering critical thinking not only enhances solution accuracy but also promotes a more robust and resilient mastery~\cite{hixon1993does,marcovitch2008self}.

Generate\&Rank~\cite{shen2021generate} is the first framework that enables models to self-evaluate their own answers. It starts with generating multiple candidate solutions for a given problem, simulating a brainstorming process to explore various possibilities. Then, a ranking module evaluates these solutions, assigning higher scores to those more likely to be correct. Both tasks are trained simultaneously by minimizing:
\begin{equation}\label{generate_rank}
    Loss=Loss^{gen}_\theta + Loss^{rank}_\theta
\end{equation}
where $\theta$ represents framework parameters. To enhance the ranker's ability to distinguish correct expressions from incorrect ones, the authors construct an expression bank to provide diverse training examples for the ranker, which employs two strategies: model-based generation and tree-based disturbance. The former relies on the generator to produce top-K expressions, each labeled as positive or negative based on whether the calculation matches the ground-truth answer. The later introduces systematic modifications to the abstract syntax tree (AST)~\cite{liu-etal-2019-tree} of the ground-truth expression. In order to make the model learn with more informative and fresh examples, the expression bank is updated online at each training epoch, ensuring the ranker is continuously exposed to a varied set of examples.

Another line to achieve critical thinking is to expose the model to multiple solutions for a given problem~\cite{webb2023emergent,sunstein1993analogical,vosniadou1989similarity}. This helps the model grasp the core reasoning path through diverse comparison and examination, rather than simply mastering some superficial patterns~\cite{patel2021nlp,gaur2023reasoning}. For this purpose, Liang et al.~\cite{liang2023generalizing} generated multiple solutions using beam search after each training iteration. These solutions are then evaluated and weighted according to their alignment with the problem to guide subsequent training and promote reasoning diversity. Zhou et al.~\cite{zhou2023learning} proposed DQGF that further expanded the expression set. It first relies on a Diverse Equations Generator to generate multiple equations (i.e., expressions) $E^{\prime}$ based on the original equation $E$, such as extracting sub-equation from $E$ or combining numbers based on their units. Then, it adopts an Equation-aware Question Generator to create a problem for each $e\in E^{\prime}$ by combining the equation with the scenario description of the original problem (tags that come with the dataset). Finally, the authors used existing powerful MWP solvers to implement a Data Filter that removes any unreasonable generated problems, ensuring only high-quality examples are retained in the augmented dataset.

\subsection{Ability 5: Knowledge Learning}
Knowledge forms the foundation of human cognition~\cite{steup2005epistemology,sloman2021cognitive}. It represents our distilled and generalized understanding of real world, serving as a crucial and indispensable scaffold for providing guidance to solve complex problems~\cite{liu2023guiding,hospedales2021meta,guo2021context}. Unlike static reliance on external knowledge bases, humans continuously acquire and internalize knowledge through experience. This dynamic relationship means that knowledge learning and mathematical reasoning should reinforce one another, creating a virtuous cycle that contributes to the growth of model intelligence~\cite{liu2023learning}. Therefore, enabling models to acquire and apply knowledge effectively is a critical step toward constructing human-like AI.

To this end, Liu et al.~\cite{liu2022cognitive} introduced CogSolver, which is the first model capable of autonomous knowledge learning. Inspired by the dual-process theory~\cite{kahneman2011thinking,evans2008dual,lieto2017dual}, CogSolver adopts a \emph{BRAIN-ARM} dual-system as its foundational structure, designed to simulate the System 1 and System 2 in human cognition, respectively. Specifically, the \emph{BRAIN} system emulates System 1 by rapidly and heuristically supplying relevant knowledge to support the reasoning process required by System 2. The \emph{ARM} system, acting as System 2, performs slow, analytical reasoning to iteratively deduce the expression. Within the dual-system framework, Liu et al. further modeled the knowledge learning process as a cycle of ``Knowledge \emph{Store-Apply-Update}'', drawing from information processing theory~\cite{atkinson1968human,ccelikoz2019cognitive,simon1978information}. In this cycle, the \emph{BRAIN} system first stores three types of knowledge: semantic knowledge, relational knowledge, and mathematical rule knowledge. Next, the \emph{ARM} system applies this stored knowledge to solve specific math word problems. Based on the feedback from problem-solving outcomes, the \emph{BRAIN} system updates its stored knowledge, refining it to enhance future reasoning tasks. Benefiting from this cycle, CogSolver autonomously accumulates knowledge through repeated problem-solving experiences, gradually forming a structured and interpretable knowledge base. This continuous refinement of knowledge also leads to sustained improvements in \emph{ARM}'s problem-solving performance.

Considering that knowledge learning is a general capability among humans~\cite{mowrer1960learning,muhajirah2020basic}, Liu et al.~\cite{liu2023learning} further introduced a LeAp framework to empower various existing MWP solvers with this ability. The core idea behind LeAp is that knowledge learning can be viewed as an encoding process from data to knowledge, while mathematical reasoning can be seen as a decoding process from knowledge to data. Based on this, LeAp diverges from traditional methods that follow a simple ``problem-expression'' translation paradigm, instead introducing a novel ``problem-knowledge-expression'' learning paradigm that explicitly incorporates knowledge as a bridge between the problems and the expressions. In implementation, LeAp employs a variational autoencoder (VAE) structure~\cite{kingma2013auto}, where a Knowledge Encoder extracts knowledge $Z$ from a set of problems $X$. A Knowledge Decoder then leverages both $X$ and $Z$ to reason out the target expressions $Y$. Existing MWP solvers can be used as the backbone of the Knowledge Decoder. In addition, the authors theoretically proved that LeAp's knowledge-learning mechanism yields more accurate results compared to traditional graph link prediction methods, making LeAp a well-founded framework.

\section{Large Language Models}
Large Language Models (LLMs) have demonstrated remarkable effectiveness across various tasks, including text comprehension~\cite{dai2023llm}, dialogue generation~\cite{liu2025socraticlm}, code writing~\cite{chen2023program,gao2023pal}, and mathematical reasoning~\cite{wei2022chain,yao2024tree}. Unlike traditional neural solvers that generate mathematical expressions, LLMs solve MWPs by generating a rational $R_P$, which integrates both solution steps and their natural language explanations. Within this rationale, LLMs unify multiple human cognitive abilities~\cite{mahowald2024dissociating}, such as problem understanding, logical organization, associative memory, and critical thinking discussed in Section 3. In this section, we review methods that selectively strengthen these cognitive abilities via different strategies.

\begin{figure*}[ht]
\setlength{\abovecaptionskip}{1pt}
\centering
\includegraphics[width=0.99\textwidth]{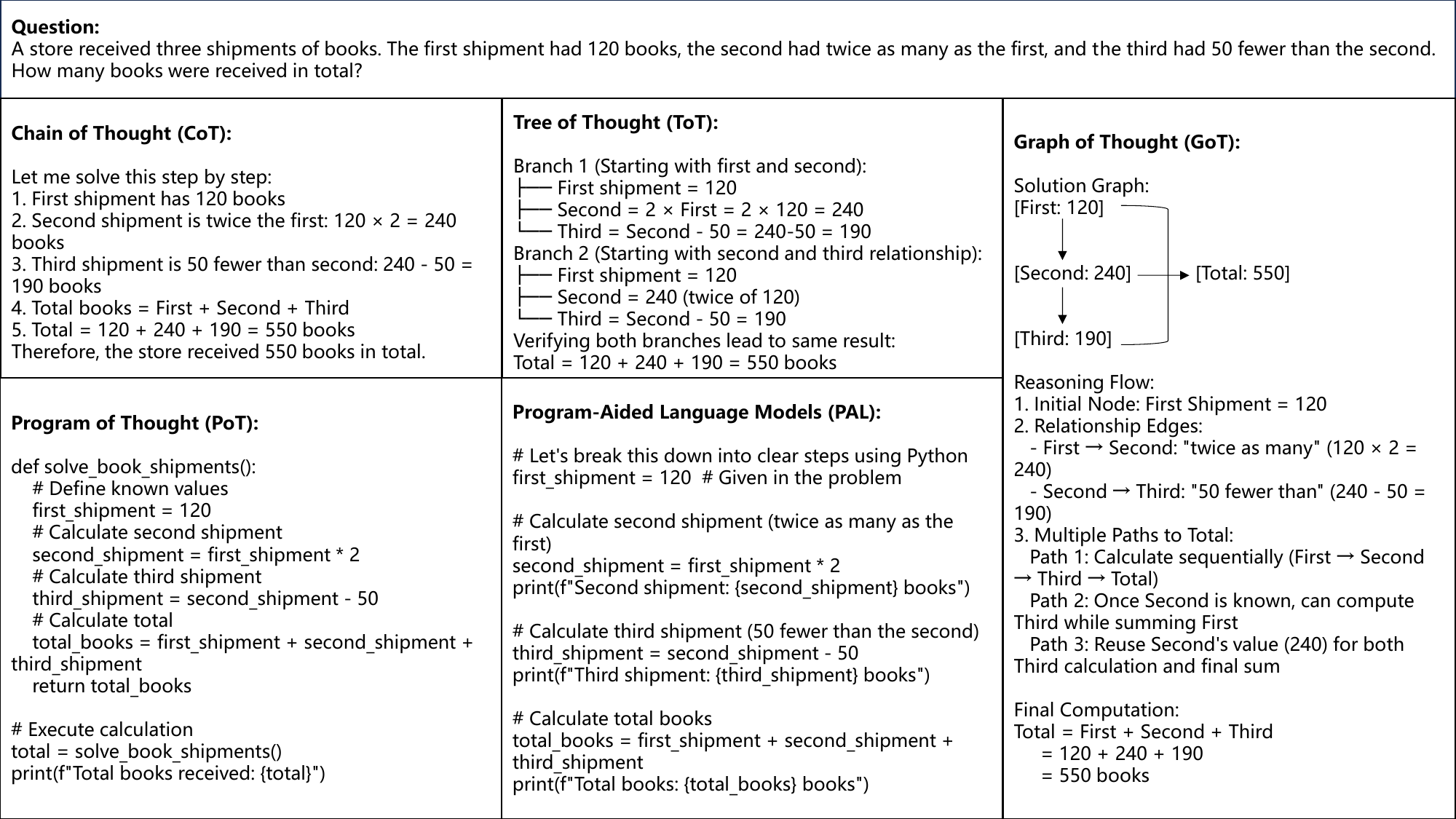}
\centering
\caption{Comparison of Different Reasoning Structures and Tool Integration Methods in LLM-based Approaches: From Sequential Reasoning (CoT) to Tree/Graph Structures (ToT/GoT) and Programming Language Integration (PoT/PAL).}
\label{fig:llm}
\end{figure*}

\subsection{General LLM-based Reasoning Methods}
Through scaling up model parameters and large-scale pre-training, LLMs have developed strong natural language understanding capabilities. This allows them to capture complex semantic relationships and numerical information in MWPs that traditional neural solvers often struggle with. In the following, we introduce several general LLM methods that focus on enhancing other cognitive abilities.

\subsubsection{Logical Organization}
When solving complex mathematical problems, humans rely heavily on logical organization - the ability to structure reasoning steps in a coherent and systematic manner~\cite{wang2010cognitive,goldin1998representational}. This cognitive capability has inspired the development of various LLM-based approaches, which have evolved from simple sequential reasoning to more sophisticated tree and graph structures. As shown in Figure~\ref{fig:llm}, Chain-of-Thought (CoT) prompting~\cite{wei2022chain,kojima2022large} marked the first significant breakthrough in this direction by enabling LLMs to break down complex problems into manageable sequential steps. Following an ``input, chain-of-thought, output'' format, this approach not only improves solution accuracy but also provides interpretable rationales that reveal the model's problem-solving process, as demonstrated in the figure where CoT clearly articulates each step: identifying the first shipment (120 books), calculating the second shipment (240 books), determining the third shipment (190 books), and finally computing the total (550 books).

While CoT successfully introduces sequential reasoning, it faces limitations with complex MWPs that require exploring multiple solution paths. For instance, in Figure~\ref{fig:llm}, when solving the bookstore problem, CoT generates a linear chain of thoughts (120→240→190→550) but struggles if an early step leads to an incorrect path, lacking the ability to systematically explore alternatives. This limitation becomes particularly evident when dealing with problems that could benefit from multiple approaches or require backtracking. To this end, Tree-of-Thoughts (ToT)~\cite{yao2024tree} advances by organizing the reasoning process into a tree structure, mirroring humans' divide-and-conquer approach to problem-solving. As demonstrated in our example, ToT explores multiple solution branches simultaneously: one branch follows the sequential calculation (First→Second→Third), while another branch explores the relationships between consecutive shipments. Both branches ultimately verify the same result of 550 books, providing multiple validation paths that were impossible with linear CoT reasoning. By maintaining multiple reasoning states and employing systematic search strategies, ToT enables LLMs to evaluate different paths, conduct look-ahead planning, and backtrack when necessary - capabilities that were missing in both traditional neural networks and basic CoT approaches.

Taking the structural evolution one step further, Graph-of-Thoughts (GoT)~\cite{besta2024graph} enhances the reasoning framework by organizing thoughts in a more flexible graph format, similar to how DAG-based neural networks improved upon tree-based approaches. As shown in Figure~\ref{fig:llm}, GoT represents the problem using three main nodes (First:120, Second:240, Third:190) connected by relationship edges (``twice as many'', ``50 fewer than''), allowing for flexible access to intermediate results. This graph structure enables multiple computation paths to reach the total of 550 books, with the ability to reuse intermediate values (like the second shipment's 240) across different reasoning paths. This framework allows for more flexible connections between reasoning steps and sophisticated backtracking strategies, particularly beneficial for problems where intermediate results can be reused across different solution paths.

Beyond predefined structures, some approaches offer even greater flexibility in logical organization. For instance, Least-to-Most~\cite{zhouleast} guides the model to decompose a problem into a sequence of subproblems and solve them incrementally. DeAR~\cite{xue2024decompose} adopts a \emph{Decompose-Analyze-Rethink} paradigm, where the model can freely update its reasoning across all prior steps at any stage. This adaptive structure maintains logical consistency throughout the reasoning process, allowing for a more structured yet flexible approach to complex problem-solving. 

\subsubsection{Associative Memory}
Associative memory enables LLMs to retrieve and integrate relevant examples or knowledge for reasoning. Among various techniques designed to enhance this capability, In-Context Learning (ICL) and Retrieval-Augmented Generation (RAG) are the most representative~\cite{coda2023meta,wang2023large}. 

ICL focuses on making LLMs learn and follow specific reasoning logical patterns based on existing examples~\cite{dong2022survey}. Its key advantage lies in the adaptability across different tasks and scenarios. However, the selection of examples remain a central challenge. To this end, the most prominent method Similar-ICL~\cite{liu2022makes,luo2023dr,fu2022complexity} aims to find examples in memory base with closest semantic representations to the test sample. The semantic representation approaches include TF-IDF, BM25~\cite{robertson2009probabilistic}, T5 encoding~\cite{raffel2020exploring}, BGE-M3~\cite{chen2024bge}, OpenAI embedding, etc. The second line of methods focus on estimating each example's influence on the test sample~\cite{peng2024revisiting}. Impact calculation approaches include influence function~\cite{van2024context,nguyen2023context}, mutual information~\cite{sorensen2022information}, perplexity~\cite{gonen2023demystifying}, code-length~\cite{wu2023self}, etc. The third category leverages LLM feedback to dynamically refine example selections~\cite{qin2024context}, adapting demonstrations based on model responses for improved reasoning performance. Notably, some studies have theoretically analyzed the impact of ICL on test performance and developed methods based on these insights. For example, Liu et al.~\cite{liumakes} proved that inference loss can be bounded by a LLM-oriented semantic similarity and the stability of the examples. Based on this, they proposed the LMS3 method, which achieved state-of-the-art results across one-shot to four-shot settings.

RAG enhances text generation by integrating a retrieval mechanism, allowing LLMs to get access to up-to-date and domain-specific external knowledge~\cite{zhang2023retrieve,fan2024survey,gao2023retrieval,huang2024survey}. The process begins with a retriever, which searches a database for relevant documents using methods like BM25 or BERT-based embeddings. The retrieved documents then serve as context for the generator, which produces outputs based on the extracted information. However, challenges such as retrieving irrelevant documents (due to poor query formulation or semantic mismatches) and incorporating noisy information (where retrieved content is incomplete, inconsistent, or redundant) can impact output quality. To mitigate these issues, techniques like query rewriting~\cite{ma2023query,mao2024rafe} refine search terms for better retrieval, while iterative retrieval-generation pipelines~\cite{zhang2023repocoder,cheng2023lift} allow models to re-evaluate and refine their outputs through multiple retrieval steps. Similar to ICL, RAG achieves strong task adaptation with minimal—if any—additional training, making it particularly effective for tasks requiring accurate and intensive knowledge, such as open-domain question answering~\cite{lewis2020retrieval}. Overall, RAG bridges the gap between internal model knowledge and external information, providing a robust framework for improving generative models.

\subsubsection{Critical thinking}
Beyond basic problem-solving, LLMs have shown remarkable capabilities in critical thinking - a cognitive skill that traditional approaches often struggle to implement. Unlike earlier NN-based methods, which relied on explicit ranking modules and carefully constructed training samples, LLMs can perform more flexible and human-like critical thinking through natural language processing~\cite{goucritic,lan2024criticeval}.

\noindent\textbf{Self-evaluation.} The first category of methods focuses on evaluating the quality of solutions before accepting them as final answers. Self-Consistency~\cite{wang2022self} adopts a simple but effective approach by generating multiple reasoning paths and using majority voting, similar to how humans verify their answers through different solution methods. Taking this evaluation process further, Self-Verification~\cite{weng2022large} introduces a more sophisticated verification mechanism that examines each reasoning step through reverse verification, particularly addressing the challenge of error accumulation in multi-step reasoning that plagues simpler approaches.

\noindent\textbf{Self-correction.} Beyond mere evaluation, more advanced methods enable LLMs to actively revise their reasoning process. The self-reflection mechanism~\cite{renze2024self} implements a structured framework of action, evaluation, and reflection, allowing LLMs to learn from past mistakes and optimize their reasoning strategies. CRITIC~\cite{gou2024critic} takes this a step further by incorporating external tool into the correction process, enabling LLMs to not only identify errors but also refine their reasoning through computational verification. This represents a more comprehensive approach to error correction, combining the flexibility of natural language reasoning with the precision of external tools.

\subsubsection{Tool Integration}
One ability that is unattainable with traditional neural networks but significantly enhances the performance of LLMs is tool utilization. It enables LLMs to overcome the inherent limitations of relying totally on pure language texts~\cite{qintoolllm,wangvoyager,ma2025advancing}. For example, despite the strong reasoning performances of CoT, it frequently makes computational errors that can critically impact the final results of MWPs.

To integrate with this cognitive capability, Program-of-Thoughts (PoT) ~\cite{chen2023program} and Program-Aided Language Models (PAL) ~\cite{gao2023pal} empower LLMs to leverage programming languages as external tools. From Figure~\ref{fig:llm}, both approaches translate mathematical reasoning into executable Python code, but with distinct styles. PoT encapsulates the entire solution in a function (e.g., $solve\_book\_shipments()$), systematically calculating the shipment values and returning the final total. In comparison, PAL adopts a more interactive approach with step-by-step execution and explicit output statements, making the reasoning process more transparent by printing intermediate results at each step.

These approaches separate the reasoning process from computation, allowing for more flexible and reliable mathematical operations. For instance, they can utilize Python's rich ecosystem of mathematical libraries - sympy for symbolic mathematics and equation solving, math for advanced arithmetic operations like square roots and logarithms, and numpy for efficient numerical computations. As shown in the example, both methods maintain computational accuracy through programmatic execution while providing different levels of granularity in their solution presentation: PoT's concise functional approach versus PAL's detailed step-by-step computation. This tool integration not only ensures computational accuracy but also provides a more versatile framework for handling diverse mathematical operations compared to the rigid expression generation of traditional methods.

Beyond utilizing existing tools, which are often limited in time and scope, the creation of new tools is equally vital. To this end, many efforts such as CREATOR~\cite{qian2023creator}, CRAFT~\cite{yuancraft}, TROVE~\cite{wang2024trove}, and KTCE~\cite{ma2024automated}, have been made to enable LLMs to generate and refine tools autonomously. For example, KTCE extracts structured mathematical knowledge from problem solutions and transforms it into reusable Python-based tools. Additionally, by simulating evolutionary operations ``selection'', ``crossover'', and ``mutation'', KTCE continuously discovers, creates, and refines new tools, increasing tool diversity. This ability further broadens the functional scope of LLMs, making them more adaptable and efficient across diverse application scenarios.

\subsection{Specialized LLM-based Reasoning Methods}
While general LLM methods showcase the unification of multiple cognitive abilities, researchers have developed specialized approaches tailored to MWP-specific challenges. These approaches address two key aspects: enhancing the interpretability of solving processes for traditional neural solvers, and specially strengthening LLMs' mathematical reasoning capabilities. This section explores how researchers leverage LLMs' unique characteristics to create more effective MWP solving systems.

\subsubsection{LLM-based MWP solvers}
This line of methods aim to combine the strengths of LLMs and NN-based models.

First, existing NN-based models are better suited for operation-level generation, such as accurately predicting the next operation in an expression, while LLMs excel at natural language interpretation. Based on this observation, Planning-LM~\cite{zhang-etal-2023-interpretable} introduces a two-stage framework. At the $t$-th reasoning step, a T5-based model is first used to encode the previous steps and, in combination with a classifier, predict the next operation, such as an ``$n+n$'' type operation. Then, the LLM generates a natural language expression of the step based on this operation. Through this iterative process, both the accuracy and interpretability of the solution are significantly improved.

Second, inspired by cognitive science research on human dual-process thinking, CoRe~\cite{zhu-etal-2023-solving} advances this direction by implementing a System 1 (quick, intuitive) and System 2 (deliberate, analytical) architecture. The framework employs a LM-based generator for rapid solution proposal and two small verification modules for rigorous evaluation, addressing the tendency of existing methods to over-rely on immediate responses without sufficient verification. Experimental results on the GSM8K dataset demonstrate CoRe's effectiveness, achieving 78.2\% accuracy and outperforming existing methods like Chain-of-Thought (63.1\%) and Self-Consistency (74.4\%) by significant margins.

Moving beyond standalone solvers, CEMAL~\cite{liang-etal-2023-gpt} explores how LLMs enhance the efficiency of neural MWP solvers through knowledge distillation. Using GTS model~\cite{xie2019goal} as the backbone, which generates expressions following a goal-decomposition process, CEMAL employs LLMs as adaptive tutors that generate customized exercises based on the student model's learning needs. This novel approach enables more efficient knowledge transfer while maintaining strong problem-solving performance, demonstrating how LLMs can be leveraged to develop more practical and scalable MWP solving systems.
\begin{table*}[t]
\small
\centering
\caption{Statistics of MWP Datasets}

\begin{tabular}{c|c|c|c|c|c|c|c}
\toprule
Dataset & \# problems & \# single-op & \# multi-op & operators & Avg. problem length & Avg. expr length & language \\
\hline
Math23K & 23162 & 4724 & 18373 & $\{+, -, \times, \div, \text{pow} \}$ & 27.9843 & 6.7949 & Chinese \\
\hline
MAWPS & 2373 & 1310 & 1062 & $\{+, -, \times, \div\}$ & 31.0147 & 6.1656 & English \\
\hline
SVAMP & 1000 & 762 & 237 & $\{+, -, \times, \div\}$ & 33.4310 & 5.9440 & English \\
\hline
MathQA & 37259 & - & - & - & 37.9 & - & English \\
\hline
GSM8K & 8792 & - & - & - & 51.79 & - & English \\
\bottomrule
\end{tabular}
\small
\label{datasets_statistics}
\end{table*}
\subsubsection{Math LLMs}
Recent advances in mathematical LLMs have demonstrated impressive reasoning capabilities across various benchmarks~\cite{sunsurvey,hong2024advances,yuemammoth,DBLP:conf/iclr/GouSGSYHDC24}. They typically rely on constructing additional data to fine-tune general LLMs. Based on the correspondence with human abilities, we provide a detailed introduction to their data construction processes.

To strengthen problem understanding ability, several approaches focus on systematic problem transformation. WizardMath~\cite{luo2023wizardmath} employs the Evol-Instruct method to vary numerical values and contextual elements while maintaining the core mathematical structure, helping models learn to identify key mathematical relationships regardless of surface variations. Similarly, Xwin-Math~\cite{xwin-lm} leverages GPT-4-Turbo to generate diverse problem variations, enabling models to develop robust understanding across different presentations of the same mathematical concepts.

Other approaches aim to enhance critical thinking through multi-perspective reasoning. MetaMath~\cite{yu2023metamath} introduces a comprehensive bootstrapping strategy that examines problems from multiple angles - forward reasoning through problem reconstruction, and backward reasoning through self-verification. This approach mirrors how humans develop deeper understanding by considering problems from different perspectives and verifying solutions through multiple paths.

To improve computational reliability, researchers have developed methods to enhance LLMs' ability to utilize external tools. Early work by Cobbe et al.~\cite{cobbe2021trainingverifierssolvemath} introduced a calculator annotation mechanism that teaches models to delegate precise calculations to an external calculator, similar to how humans might use calculators for complex arithmetic. Through special formatting (e.g., ``$5\times4=\ll5\times4=20\gg$''), the model learns to interface with the calculator during inference, significantly reducing computational errors while maintaining its natural language reasoning capabilities.
\begin{table*}[h!]
\centering
\caption{Performances of NN-based Methods on MWP datasets}
\fontsize{9pt}{14pt}\selectfont
\begin{tabular}{c|c|c|c|c|c}
\toprule
 & \textbf{Method} & \textbf{Math23K} & \textbf{MAWPS} & \textbf{SVAMP} & \textbf{MathQA} \\ \toprule
\multirow{6}{*}{\textbf{Problem Understanding}} 
  & DNS~\cite{wang2017deep}             &        58.1    &       59.2     &     20.0       &    /        \\
  & HMS~\cite{lin2021hms}             &    76.1        &     80.3       &     17.9      &       /     \\
  & Graph2Tree~\cite{zhang2020graph}      &     77.4       &      83.7      &    31.9        &      69.5     \\
  & KA-S2T~\cite{wu2020knowledge}          &     76.3       &     /       &      /      &      /      \\
  & MWP-BERT~\cite{liang2022mwp}        &    84.7        &     82.9       &     /       &     76.2       \\ 
  & BERT-Tree~\cite{li2022seeking}        &    83.3       &      87.2     &      32.4      &       73.8     \\ \hline
\multirow{3}{*}{\textbf{Logical Organization}} 
  & GTS~\cite{xie2019goal}             &   74.3         &    82.6        &    27.7      &      /      \\
  & FOMAS~\cite{liu2023guiding}           &   84.8         &    88.6        &   /         &      /      \\
  & DEDUCTREASONER~\cite{jie2022learning}  &   85.1         &    92.0        &   47.3       &        78.6   \\ \hline
\multirow{2}{*}{\textbf{Associative Memory}}  
  & RHMS~\cite{lin2023learning}           &    78.6        &      84.9      &     /       &      /      \\ 
  & REAL~\cite{huang2021recall}           & 82.3               &        /        &      /      &       /     \\ \hline
\multirow{1}{*}{\textbf{Critical Thinking}}   
  & Generate\&Rank~\cite{shen2021generate}  &   85.4        &       84.0     &       /     &        /    \\ \hline
\multirow{2}{*}{\textbf{Knowledge Learning}}  
  & CogSolver~\cite{liu2022cognitive}       &    77.3        &      82.9      &        /    &      /      \\ 
  & LeAp~\cite{liu2023learning}         & 77.9          &    85.2          &    34.1        &   /         \\ 
\bottomrule
\end{tabular}
\label{nn_results}
\end{table*}
\section{Dataset \& experiment}
In the realm of math word problems, various methods are evaluated on datasets which are usually collected from educational resources and expert-annotated problem sets. In this section, we delve into a comprehensive overview of the datasets widely utilized in existing studies, discussing their characteristics and the contexts in which they are applied. Following this, we analyze the performances of different methods tested on these datasets. 
By collecting results into two tables, we aim to provide a clear comparison among these methods, facilitating a more intuitive understanding of current advancements in this field.

\subsection{Datasets}
Several datasets have been proposed for MWPs, starting with early efforts such as Dolphin18K~\cite{huang2016well}, ALG514~\cite{kushman2014learning}, and DRAW1K~\cite{upadhyay2016learning}. At present, the most frequently used benchmarks include Math23K, MAWPS, SVAMP, MathQA, and GSM8K. We introduce these five datasets below and report their statistics in Table~\ref{datasets_statistics}.

\begin{enumerate}
\item{Math23k~\cite{wang2017deep} comprises 23,162 elementary math problems sourced from diverse educational websites. Each problem involves one unknown variable. The dataset was constructed from an initial pool of about 60,000 problems, where equation templates were first extracted using a rule-based approach and then filtered to ensure quality.}
\item{MAWPS~\cite{koncel2016mawps} is an online repository that unifies datasets from several previous works~\cite{hosseini2014learning, kushman2014learning, koncel2015parsing, roy2015solving}. The online nature of this repository allows for the automatic construction of datasets with particular characteristics from web-sourced corpora. In practice, most studies use a standardized subset of 2,373 single-variable problems.}
\item{SVAMP~\cite{patel2021nlp} is a challenging dataset of 1,000 single-variable arithmetic word problems of grade level up to 4. The problems in the dataset are created by applying certain variations to existing problems, aimed at evaluating model robustness.}
\item{MathQA~\cite{amini2019mathqa,ling2017program} is a large-scale dataset comprising 37,200 problems. Each problem is provided with a question text, a list of multiple-choice options, the correct answer, and an aligned operation program. The problems in the dataset come from the AQuA dataset, carefully annotated with a representation language and annotation system that addresses unwanted noise in the dataset and lack of formal operation-based representations.}
\item{GSM8K~\cite{cobbe2021training} comprises 8,792 grade-school problems carefully crafted by human authors to ensure linguistic diversity and high quality. The dataset is split into 7,473 training problems and 1,319 test problems. Each problem is provided with a question text and a solution requiring 2 to 8 steps, typically involving a sequence of elementary arithmetic operations ${+, -, \times, \div }$. This structure makes GSM8K a valuable resource for training models in multi-step mathematical reasoning.}
\end{enumerate}

\subsection{Experimental Results}
{\color{black}{Based on the datasets mentioned, we conduct experiments to investigate the performances of different methods and present the results in Table~\ref{nn_results} and Table~\ref{llm_results}, which list representative NN-based methods and LLM-based methods, respectively. To ensure fairness, all experiments were consistently conducted on the same benchmark datasets under identical evaluation methods. We maintain a public repository and consistently update resources\footnote{https://github.com/ShangziXue/MWPSurvey}.}}

Each method in Table~\ref{nn_results} and Table~\ref{llm_results} is categorized according to the abilities described in above sections. This organization allows readers to easily compare the performance of various approaches across the datasets. The entries in the tables display the accuracies achieved by these methods, with cells marked as ``-'' indicating the absence of experimental data for a particular method on a given dataset. From the data, we extract notable observations and provide explanations for the outcomes.
\begin{table*}[h!]
\centering
\caption{Performances of LLM-based Methods on MWP datasets}
\fontsize{9pt}{14pt}\selectfont
\begin{tabular}{c|c|c|c|c|c|c|c}
\toprule
 & \textbf{Method} & \textbf{Backbone} & \textbf{Math23K} & \textbf{MAWPS} & \textbf{SVAMP} & \textbf{MathQA} & \textbf{GSM8K} \\ 
\toprule
\multirow{6}{*}{\textbf{Logical Organization}} 
  & \multirow{2}{*}{CoT~\cite{wei2022chain}}              & GPT-3.5           &  85.8        &  91.9         &  88.4        &  79.8        &   87.2       \\
  &                  & LLaMA3.1-8B       &    83.4       &   92.7      &   88.9      &  80.2        &  87.4         \\ \cline{2-8}
  & \multirow{2}{*}{ToT~\cite{yao2024tree}}              & GPT-3.5           &  86.5          &  92.5        &    89.7       &    80.8        &   88.8        \\
  &                  & LLaMA3.1-8B       &   84.9      &  93.2        &  90.0       & 81.0    &  89.2         \\ \cline{2-8}
  & \multirow{2}{*}{GoT~\cite{besta2024graph}}              & GPT-3.5           &  86.0          &  93.0     & 90.1        & 80.6       &  87.9       \\
  &                  & LLaMA3.1-8B       &  84.4          &   93.9         & 90.8          &  81.0          &  88.5        \\ \hline
\multirow{2}{*}{\textbf{Associative Memory}}  
  & \multirow{2}{*}{ICL~\cite{dong2022survey}}              & GPT-3.5         &   85.9       &  92.8       &  89.9        & 81.1      & 90.8     \\
  &                  & LLaMA3.1-8B       &  83.7        &   93.3         &   91.0         & 81.5         &   91.3          \\ \hline
\multirow{4}{*}{\textbf{Critical Thinking}} 
  & \multirow{2}{*}{Self-Consistency~\cite{wang2022self}} & GPT-3.5           &   87.0        &  93.1        &  91.7    & 81.4     &  90.5         \\
  &                  & LLaMA3.1-8B       &  84.1        &   94.5         &  91.9         &   81.7         &  91.2         \\ \cline{2-8}
  & \multirow{2}{*}{Self-Verification~\cite{weng2022large}}  & GPT-3.5         &  86.8          &   94.4      & 90.6       & 82.5       &  91.6    \\
  &                  & LLaMA3.1-8B       &  84.6       &  93.8       & 92.1      &  82.0    &  90.9       \\ \hline
\multirow{4}{*}{\textbf{Tool Integration}} 
  & \multirow{2}{*}{PoT~\cite{chen2023program}}              & GPT-3.5        &  88.3          &  95.0      & 93.8         &  84.7       &  92.8        \\
  &                  & LLaMA3.1-8B       &  85.8      &   95.8        & 93.5         & 85.5          & 92.5           \\ \cline{2-8}
  & \multirow{2}{*}{PAL~\cite{gao2023pal}}              & GPT-3.5           & 87.7          &  95.3         &   92.5       &   83.1       &    93.0        \\
  &                  & LLaMA3.1-8B       &  86.1        &   96.0         &   93.7        &  82.4         &    93.4        \\ \hline
\multirow{4}{*}{\textbf{Math LLMs}} 
  & \multirow{2}{*}{WizardMath~\cite{luo2023wizardmath}}       & LLaMA2-7B         &  75.2         &  79.5       &  63.2          &  73.5          &  75.1        \\ 
  &                  & LLaMA2-70B        &  85.8          & 88.6           &   76.4         &  80.1          &   83.8           \\ \cline{2-8}
  & \multirow{2}{*}{MetaMath~\cite{yu2023metamath}}         & LLaMA2-7B         &  74.4         &  82.4          &  75.8         &   77.6        &  79.2      \\ 
  &                  & LLaMA2-70B        &  84.5          &   89.3         &  80.6          &    81.0        &  85.3          \\

\bottomrule
\end{tabular}
\label{llm_results}
\end{table*}

For NN-based methods, we observe the following insights from the experimental results in Table~\ref{nn_results}: (1) On average, methods categorized under abilities such as ``Logical Organization'' and ``Critical Thinking'' achieve higher accuracy compared to those focused solely on ``Problem Understanding''. This indicates that while problem understanding is a fundamental skill in mathematical reasoning, the improvements in model performance are more pronounced with the incorporation of ``Logical Organization'' and ``Critical Thinking''. (2) Among ``Problem Understanding'' methods, approaches that utilize more sophisticated tree-structured representations, such as HMS, Graph2Tree, and KA-S2T, outperform simpler seq2seq-based models like DNS. This demonstrates that finer-grained modeling of problem features can significantly improve problem understanding. (3) Within ``Logical Organization'' methods, other approaches outperform the basic goal-driven organization method (i.e., GTS), with gains in this dimension being more pronounced than in other abilities. For instance, FOMAS benefits from incorporating formula knowledge, achieving a 10.5\% improvement, while DEDUCTREASONER improves performance by 10.8\% through scoring sub-expressions and updating operand representations. These enhancements demonstrate that improvements in ``Logical Organization'' lead to the most significant gains in reasoning accuracy. 

For LLM-based methods, we use GPT-3.5 and LLaMA3.1-8B as the backbone and present the results of several representative approaches in Table~\ref{llm_results}. The strong reasoning abilities of LLMs enable them to outperform NN-based approaches on most of the datasets. Furthermore, we observe that (1) Among ``Logical Organization'' methods, with the same backbone LLM, ToT and GoT consistently outperform CoT, which suggests that mimicking human logical thinking by allowing LLMs to explore multiple reasoning paths and self-evaluate choices is effective. 
(2) ``Tool Integration'' methods, such as PoT and PAL, outperform ``Logical Organization'' methods. These approaches utilize LLMs to generate programming language statements and execute them using external tools, demonstrating that generating code and leveraging tool-based execution offers a more effective enhancement for mathematical reasoning tasks compared to purely logical organization in natural language. (3) In-context Learning (ICL) surpasses CoT, highlighting the effectiveness of the associative memory mechanism, which mimics human cognition by using task-relevant examples to guide reasoning. (4) ``Critical Thinking'' methods outperform ``Logical Organization'' methods across most of the datasets, underscoring the significant role of enabling LLMs to evaluate, reflect on, and correct their own outputs. This emphasizes the importance of self-assessment and refinement capabilities in enhancing mathematical reasoning performance.

\section{Mathematical Tasks Beyond MWP}
Beyond MWPs, there exist many other critical mathematical reasoning tasks that require more complex cognitive skills~\cite{xiao2024learning,loveland2016automated}. For example, multi-modal perception ability is essential for geometry problem solving~\cite{xiao2024learning}, as it involves interpreting geometric diagrams alongside textual descriptions. Similarly, strategic planning and symbolic computation abilities play a crucial role in automatic theorem proving~\cite{loveland2016automated}, which focuses not on numerical calculation but on rigorously proving theorems through logical reasoning. Additionally, advanced mathematical domains such as calculus, algebra, number theory, and probability demand increasingly sophisticated reasoning skills, challenging a broader and deeper range of human cognitive abilities to tackle these diverse and intricate problems.
\subsection{Geometry Problem Solving}
\begin{figure*}[t]
    \centering
    \subfloat[]{\includegraphics[width=0.34\textwidth]{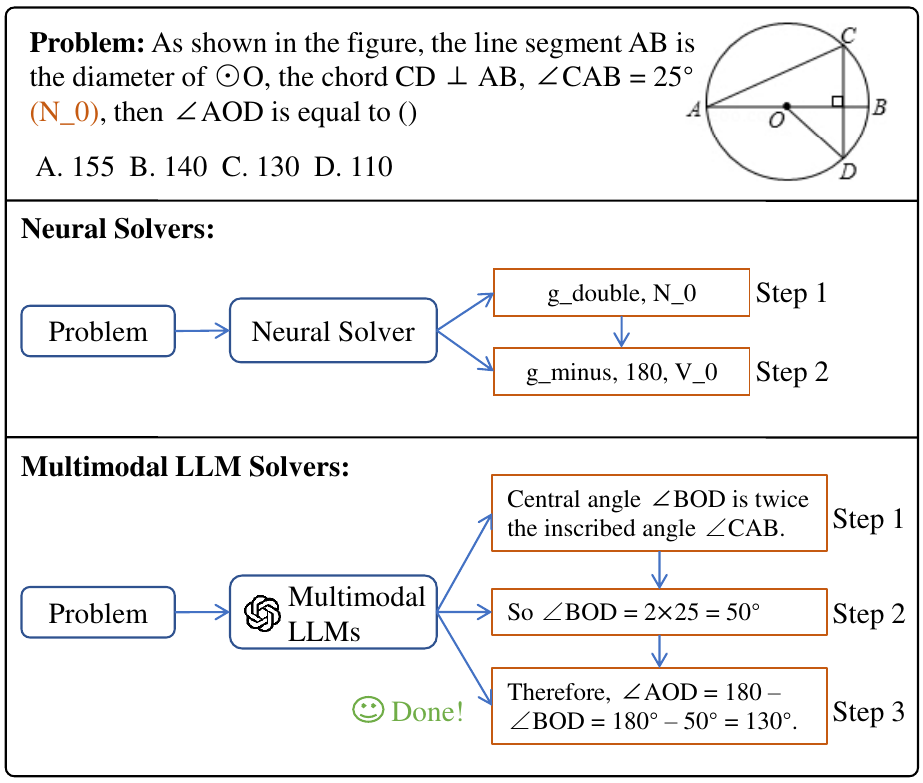} \label{fig:other-math-example-gps}} 
    \hfill
    \subfloat[]{\includegraphics[width=0.29\textwidth]{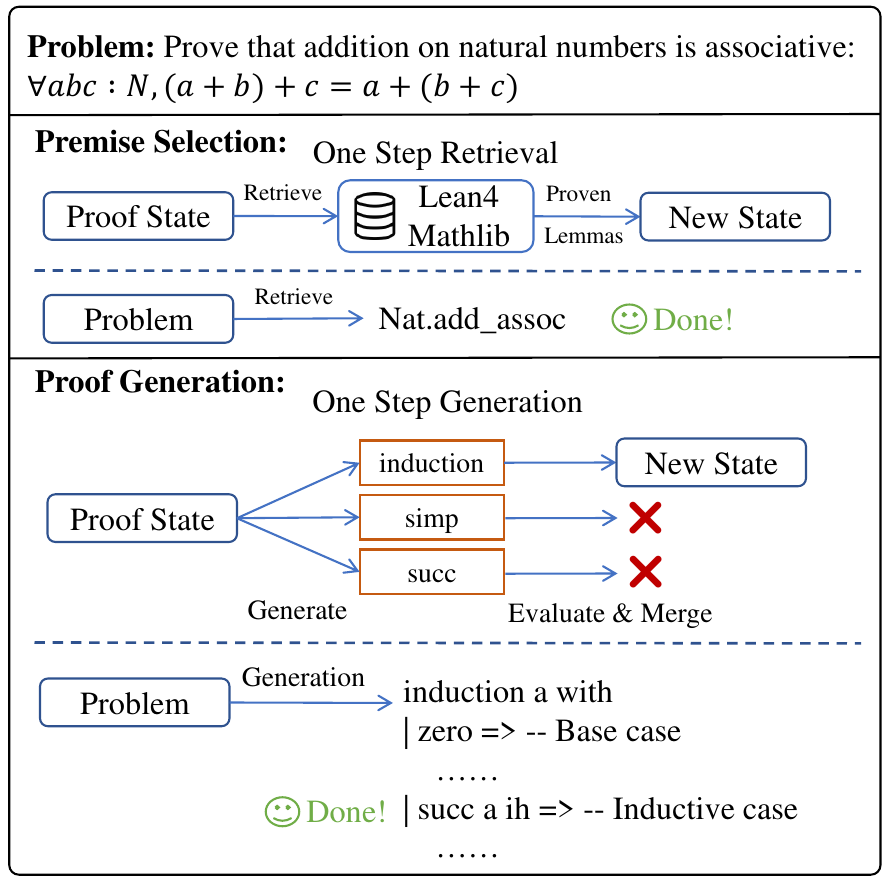} \label{fig:other-math-example-atp}} 
    \hfill
    \subfloat[]{\includegraphics[width=0.342\textwidth]{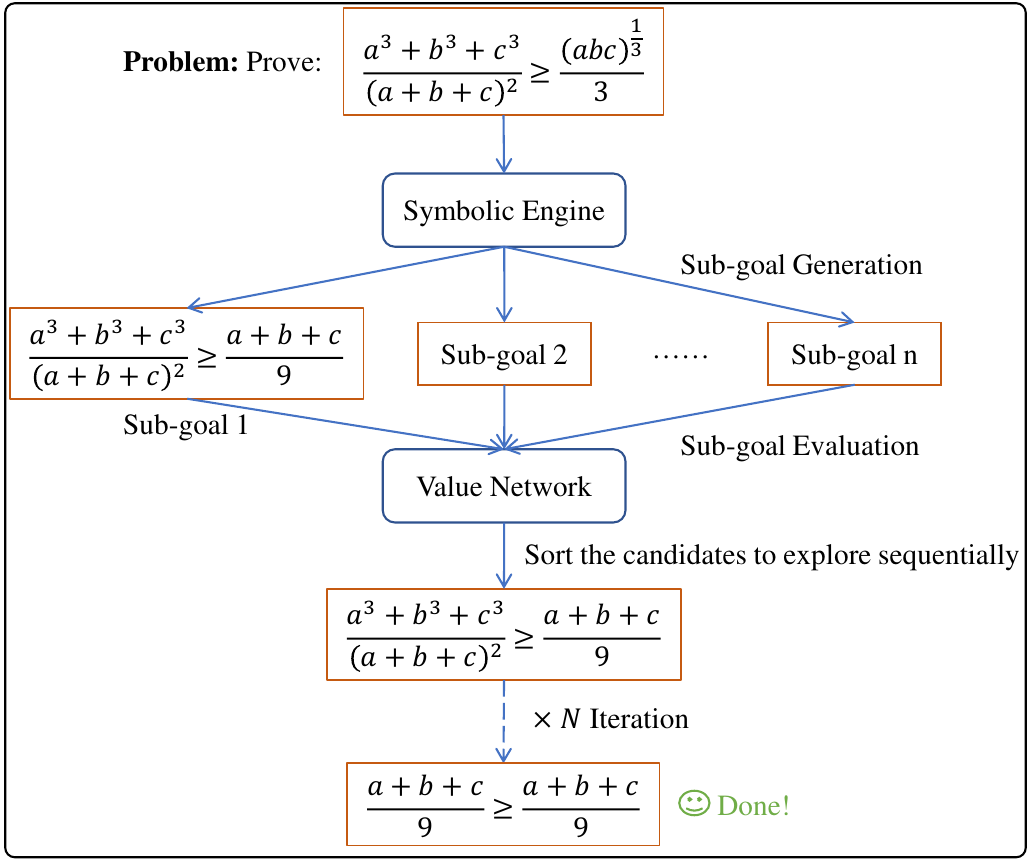} \label{fig:other-math-example-inequality}} 
    \caption{Examples of three Mathematical Tasks beyond MWP.}
    \label{fig:1}
\end{figure*}

Geometry Problem Solving (GPS) is a classic and challenging task that has gained significant attention in recent years~\cite{Seo2015SolvingGP,Seo2014DiagramUI,Lu2021InterGPSIG,Chen2021GeoQAAG,Cao2022AnAB}. As illustrated in Figure~\ref{fig:other-math-example-gps}, solving these problems requires not only multimodal understanding to interpret the information within the problem and diagram and the ability to apply geometry theorems accurately to solve problems. Current existing works on solving geometry problems primarily focus on tackling geometry problems at the elementary and middle school levels.

Early geometry problem solving methods focus on multi-modal perception. Seo et al.\cite{Seo2015SolvingGP} proposed the first symbolic solver GeoS, which first utilized handcrafted regex rules and OCR techniques~\cite{Seo2014DiagramUI} to parse the problem text and diagram into first-order logic literals. The solver then searched for an assignment that satisfied all parsed literals to determine the solution.
To further enhance human readability of the reasoning process, Inter-GPS \cite{Lu2021InterGPSIG} improves GeoS by iteratively searching geometry primitives and applying a series of manually defined geometry theorems. Although above symbolic solvers possess strong process interpretability, they heavily rely on the human handcrafted regex rules to parse the problem text and diagram, which limits their adaptability to unseen geometric data.

To address the generalization issues of symbolic solvers, deep learning based methods have emerged and been rapidly explored then, aiming to further improve the multi-modal perception abilities of geometry solvers in an end-to-end manner. Chen et al.\cite{Chen2021GeoQAAG} proposed the first neural geometry solver called NGS which solved the geometry problems with an encoder-decoder framework. It first fuses the problem text and diagram through multi-modal fusion module, and then sends the fused features to a program decoder to generate program tokens that can obtain final results through program execution, as Figure~\ref{fig:other-math-example-gps} shown.
On this basis, DPE-NGS~\cite{Cao2022AnAB} enhanced the text encoder by employing Bi-LSTM and RoBERTa~\cite{Liu2019RoBERTaAR} for encoding simultaneously, which was further improved by SCA-GPS~\cite{Ning2023ASC} through integrating diagram features with symbolic characters. Apart from enhancing the multi-modal understanding ability of the neural solver, Xiao et al.\cite{xiao2024learning} proposed DualGeoSolver, which aims to strengthen knowledge learning and geometric reasoning process by constructing a dual-process system. This system explicitly incorporates geometric knowledge into the reasoning process, significantly improving the solver’s performance on geometry problems.
Moreover, to enhance diagram interpretation, Zhang et al.\cite{Zhang2022PlaneGD} introduced PGDPNet, which leverages computer vision techniques~\cite{he2017mask,ying2021embed,xu2017scene} to parse the geometry primitives and their relationships from diagrams. Building on PGDPNet’s parsing capabilities, PGPSNet \cite{Zhang2023AMN} is subsequently developed to solve geometry problems by applying semantic embeddings to various types of geometric primitives.

The past two years have seen rapid advancements in large language models (LLMs) and multimodal LLMs (MLLMs). Given their strong generalization and generative capabilities, many recent works have explored applying LLMs and MLLMs to the field of geometry problem solving. AlphaGeometry~\cite{trinh2024solving} solved complex geometry problems at a level approaching a human Olympiad gold-medalist by adopting a neuro-symbolic system. It trained a language model on a synthetic dataset with 100 million data samples to construct auxiliary points and lines, then employed symbolic engines for deduction and reasoning. G-LLaVA~\cite{gao2023g} represents the first attempt to develop a dedicated geometry problem solver built entirely on MLLMs. Difference from neural solvers, G-LLaVA generates step-by-step rationales instead of programs, offering a more human-readable approach to solving geometry problems. It first constructed a large-scale dataset, Geo170K, by augmenting the GeoQA and Geometry3K datasets, then fine-tuned the LLaVA model on this dataset, surpassing previous methods. Recently, DFE-GPS~\cite{zhang2024diagram} constructed a new large-scale geometric dataset SynthGeo228K, which contains formal and natural language annotations and diverse geometric diagrams that were generated by proposed Geometry Model Builder. DFE-GPS then was trained on SynthGeo228K with natural and formal language supervision simultaneously, demonstrating robust multi-modal understanding and reasoning capabilities and achieving new state-of-the-art performance on FormalGeo7K benchmark.

\subsection{Automatic Theorem Proving}
Automatic Theorem Proving (ATP) is one of the core components of mathematics reasoning, with a level of logical complexity far exceeding that of MWPs and GPS~\cite{schulz2002brainiac,kovacs2013first,chou2000deductive,korovin2008iprover,barbosa2022cvc5}. Its difficulty, often reaching university-level complexity, poses significant challenges for problem understanding, logical organization and planning, and symbolic computation. Recently, advancements in LLMs have driven substantial research into leveraging these techniques to enhance the automatic theorem proving process. Existing works can be categorized into two genres: Premise Selection and Proof Generation.

\subsubsection{Premise Selection}
Premise Selection is modeled as retrieving premises from previous proven lemmas (\textit{Lean4 MathLib} in Figure~\ref{fig:other-math-example-atp}) that that will most likely lead to an automatically constructed proof~\cite{irving2016deepmath}, which requires the provers to possess strong problem understanding and logical organization abilities. As the pioneer of the ATP, \cite{irving2016deepmath} modeled premise selection as a binary classification task, which embedded each premise and the target statement through convolutional network or recurrent network jointly and then predict the probability of helpfulness of this premise to the target. Building on this framework, follow up works enhance the representation capability of the embedding extractor layers to achieve better problem understanding. \cite{piotrowski2020stateful} introduced recurrent neural networks to capture stateful information within the proof trajectory, moving beyond merely predicting pairwise information between each premise and target. \cite{prorokovic2021improving} utilized Transformer to achieve a more robust and powerful representation of premises and targets, leading to improved prediction performance. 

On the other hand, a stream of research utilized the inherent structure information inside the mathematical formulas. \cite{wang2017premise} represents higher-order logic formulas as directed acyclic graphs and embed these formula graphs into feature vectors while preserving the information of edge ordering using a novel method similar to message passing on graphs, achieving SOTA performance on HOlStep dataset at that time. \cite{paliwal2020graph} then optimized the embedding process of logic formulas in HOL Light by graph neural networks. Subsequently, \cite{lin2021contrastive} developed a graph contrastive learning method to improve the graph representation of logical formulas for premise selection. Recent years, with the rapid developments of the pre-trained language models, many works have introduced BERT-series or GPT-series language models into premise selection thanks to their superior representation ability. \cite{ferreira2020natural,Ferreira2021STARC} fine-tuned a BERT-like model to encode both logic formulas and natural language within proofs to enhance premise retrieval. Meanwhile, PACT~\cite{han2021proof} employed an auto-regressive objective for premise selection with the assistance of Lean formal language.

\subsubsection{Proof Generation}
Unlike Premise Selection, which focuses on retrieving relevant premises from a large knowledge base, proof generation aims to produce one or more steps based on the current proof state to build the entire proof, which requires more advanced logical organization and planning capabilities. Proof generation can be further divided into two categories based on the locality of the generation process: proof-step generation and proof-trajectory generation~\cite{huang2018gamepad,yang2019learning,polu2020generative,welleck2022naturalprover,jiang2022thor,zhang2023getting,yousefzadeh2023large,azerbayev2023llemma,shao2402deepseekmath,frieder2024mathematical,bansal2019holist,polu2020generative}.

GamePad~\cite{huang2018gamepad} encodes proof state using a TreeLSTM network and generates next proof step through collaboration of position evaluation and tactic prediction based on the proof state feature. Compared to GamePad that selects tactics from a fixed tactic set, ASTactic~\cite{yang2019learning} constructs tactics as programs by assembling them into abstract syntax trees (ASTs) within a structured, predefined context-free grammar. Besides, there arises many works that utilize language models to generate proof steps. GPT-\textit{f}~\cite{polu2020generative} trained a decoder-only Transformer model to generate the next proof step conditioned to a specially formatted prompt: \texttt{GOAL <GOAL> PROOFSTEP}. Then, unlike GPT-\textit{f} which generates formal proof steps expressed in programming language, NaturalProver~\cite{welleck2022naturalprover} generates natural mathematical language for proving. It fine-tunes GPT-3 to generate proofs by conditioning on background references, which are retrieved or human-provided theorems or definitions, using a constrained decoding approach. Thor~\cite{jiang2022thor} integrates premise selection into the proof step generation process of language models by introducing a \texttt{<hammer>} token, which invokes ATP tools for premise selection to simplify the proof. As LLMs continue to showcase their powerful reasoning capabilities, many works have employed LLMs to generate proof steps. \cite{zhang2023getting,yousefzadeh2023large,azerbayev2023llemma,shao2402deepseekmath,frieder2024mathematical} have explored the mathematical and theorem-proving capabilities of LLMs through both prompting techniques and fine-tuning approaches.

Proof-trajectory generation aims to explore a vast search space with complex search algorithms and heuristics to efficiently explore potential proof paths, guiding the model toward promising branches while avoiding unproductive ones. As illustrated in Figure~\ref{fig:other-math-example-atp}, they typically adopt a value network to evaluate the reward of each step, and further employ complex search algorithm (e.g., Monte Carlo Tree Search) to find the locally optimal trajectory.

Some works utilize tree traversal algorithm to search the best trajectory from search space, such as breadth-first search~\cite{bansal2019holist}, depth-first search~\cite{yang2019learning} and best-first search~\cite{polu2020generative}. Notably, GPT-\textit{f}~\cite{polu2020generative} maintains a proof search tree and cumulative logprob queue to explore and expand the goals in which the language model has the highest global confidence. Building on this intuition, many works~\cite{rawson2019neurally,lample2205hypertree,wang2023dt,brandfonbrener2024vermcts} have introduced Monte Carlo Tree Search (MCTS) into proof-trajectory search and generation. For instance, DT-Solver~\cite{wang2023dt} introduces a proof-level value function to estimate the value and confidence of each state, enhancing standard MCTS by employing dynamic tree sampling based on these estimations. Additionally, many studies~\cite{kusumoto2018automated,fawzi2019learning,wu2021tacticzero,crouse2021deep,abdelaziz2022learning,fokoue2023ensemble} have approached proof-trajectory generation as a Markov Decision Process, applying deep reinforcement learning to address it. Specifically, \cite{fawzi2019learning} designs a deep Q-network to estimate the value of proof state and action polynomial, guiding the choice of elementary inference rules for proving polynomial inequalities. TRAIL~\cite{abdelaziz2022learning} trains a attention-based policy network using policy gradient loss and entropy regularization in saturation-based provers.

\subsection{Other Mathematical Problems}
Several studies have explored the application of machine learning techniques to domains requiring a broader and deeper range of human cognitive abilities~\cite{dabelow2025symbolic,wei2024proving,he2020machine,lee2024exploring}. However, they predominantly focus on specialized, narrowly defined aspects of the field. For instance, AIPS\cite{wei2024proving} is designed to tackle complex inequality theorems at the International Mathematical Olympiad (IMO) level, utilizing a dedicated symbolic deductive engine integrated with a value network refined through curriculum learning to optimize proof search capabilities. As illustrated in Figure~\ref{fig:other-math-example-inequality}, each iteration begins with the symbolic engine generating sub-goals by applying predefined theorems. The value network then evaluates and ranks these sub-goals for the next iteration. This process continues until a trivial inequality or a predefined theorem is reached. Although these methods excel within their specialized domains, they often fall short in addressing broader challenges, particularly those requiring intricate and multifaceted problem-solving capabilities.

Recent advancements in math LLMs\cite{yu2023metamath,luo2023wizardmath,tong2024dart} have demonstrated impressive performance on the mathematical tasks such as MATH dataset~\cite{hendrycks2021measuring}, showcasing notable capabilities in domains such as algebra, number theory, and probability. However, since these math LLMs are primarily trained using general language modeling techniques, without modeling the cognitive abilities of humans, their performance improvements may stemming from the augmentation of training split of MATH dataset. Their ability to handle problems that exceed the difficulty of MATH or those not present in the dataset remains unclear. Therefore, the development of robust methodologies for addressing the full spectrum of mathematical problems within these domains is still an open and under-explored domain.

\section{Conclusion}
In this paper, we provided a comprehensive analysis of math word problem (MWP) solving approaches from a human cognitive perspective. We focused on five key cognitive abilities: \emph{Problem Understanding}, \emph{Logical Organization}, \emph{Associative Memory}, \emph{Critical Thinking}, and \emph{Knowledge Learning}. We compared existing methods based on these abilities, offered a structured classification, and conducted detailed experimental evaluations. Furthermore, we analyzed the cognitive abilities of both small models and large language models (LLMs) separately, highlighted their distinct strengths and weaknesses in solving MWP tasks, and provided valuable insights for improving reasoning accuracy in these models.

\section*{Acknowledgement}
This work was supported by the National Natural science Foundation of China (Grant No.62477044, No.U23A20319), the Key Technologies R\&D Program of Anhui Province (No.202423k09020039), the Fundamental Research Funds for the Central Universities (No.WK2150110038). Zhenya Huang gratefully acknowledges the support of the Young Elite Scientists Sponsorship Program by CAST (No. 2024QNRC001). YWT is supported by the Ministry of Digital Development and Information (MDDI) under the Singapore Global AI Visiting Professorship Program (Award No. AIVP-2024-002).
\bibliographystyle{IEEEtran}
\bibliography{bib}

\begin{thebibliography}{100}
\providecommand{\url}[1]{#1}
\csname url@samestyle\endcsname
\providecommand{\newblock}{\relax}
\providecommand{\bibinfo}[2]{#2}
\providecommand{\BIBentrySTDinterwordspacing}{\spaceskip=0pt\relax}
\providecommand{\BIBentryALTinterwordstretchfactor}{4}
\providecommand{\BIBentryALTinterwordspacing}{\spaceskip=\fontdimen2\font plus
\BIBentryALTinterwordstretchfactor\fontdimen3\font minus \fontdimen4\font\relax}
\providecommand{\BIBforeignlanguage}[2]{{%
\expandafter\ifx\csname l@#1\endcsname\relax
\typeout{** WARNING: IEEEtran.bst: No hyphenation pattern has been}%
\typeout{** loaded for the language `#1'. Using the pattern for}%
\typeout{** the default language instead.}%
\else
\language=\csname l@#1\endcsname
\fi
#2}}
\providecommand{\BIBdecl}{\relax}
\BIBdecl

\bibitem{davies2021advancing}
A.~Davies, P.~Veli{\v{c}}kovi{\'c}, Buesing \emph{et~al.}, ``Advancing mathematics by guiding human intuition with ai,'' \emph{Nature}, vol. 600, no. 7887, pp. 70--74, 2021.

\bibitem{zhang2020gap}
D.~Zhang, L.~Wang \emph{et~al.}, ``The gap of semantic parsing: A survey on automatic math word problem solvers,'' \emph{IEEE Transactions on Pattern Analysis and Machine Intelligence}, vol.~42, no.~9, pp. 2287--2305, 2020.

\bibitem{bakman2007robust}
Y.~Bakman, ``Robust understanding of word problems with extraneous information,'' \emph{arXiv preprint math/0701393}, 2007.

\bibitem{wang2017deep}
Y.~Wang, X.~Liu, and S.~Shi, ``Deep neural solver for math word problems,'' in \emph{Proceedings of the 2017 conference on empirical methods in natural language processing}, 2017, pp. 845--854.

\bibitem{wang2019template}
L.~Wang \emph{et~al.}, ``Template-based math word problem solvers with recursive neural networks,'' in \emph{AAAI}, 2019, pp. 7144--7151.

\bibitem{shi2015automatically}
S.~Shi, Y.~Wang, C.-Y. Lin, X.~Liu, and Y.~Rui, ``Automatically solving number word problems by semantic parsing and reasoning,'' in \emph{EMNLP}, 2015, pp. 1132--1142.

\bibitem{huang2018neural}
D.~Huang, J.~Liu, C.-Y. Lin, and J.~Yin, ``Neural math word problem solver with reinforcement learning,'' in \emph{Proceedings of the 27th International Conference on Computational Linguistics}, 2018, pp. 213--223.

\bibitem{shen2021generate}
J.~Shen, Y.~Yin, L.~Li, L.~Shang, X.~Jiang, M.~Zhang, and Q.~Liu, ``Generate \& rank: A multi-task framework for math word problems,'' in \emph{EMNLP}, 2021, pp. 2269--2279.

\bibitem{zhang2020graph}
J.~Zhang, L.~Wang, R.~K.-W. Lee \emph{et~al.}, ``Graph-to-tree learning for solving math word problems,'' in \emph{ACL}, 2020, pp. 3928--3937.

\bibitem{li2019modeling}
J.~Li, L.~Wang, J.~Zhang, Y.~Wang, B.~T. Dai, and D.~Zhang, ``Modeling intra-relation in math word problems with different functional multi-head attentions,'' in \emph{ACL}, 2019, pp. 6162--6167.

\bibitem{qin2021neural}
J.~Qin, X.~Liang, Y.~Hong, J.~Tang, and L.~Lin, ``Neural-symbolic solver for math word problems with auxiliary tasks,'' in \emph{ACL/IJCNLP}, 2021, pp. 5870--5881.

\bibitem{lin2021hms}
X.~Lin, Z.~Huang \emph{et~al.}, ``Hms: A hierarchical solver with dependency-enhanced understanding for math word problem,'' in \emph{AAAI}, vol.~35, no.~5, 2021, pp. 4232--4240.

\bibitem{wu2020knowledge}
Q.~Wu, Q.~Zhang, J.~Fu, and X.-J. Huang, ``A knowledge-aware sequence-to-tree network for math word problem solving,'' in \emph{EMNLP}, 2020, pp. 7137--7146.

\bibitem{yang2022logicsolver}
Z.~Yang, J.~Qin, J.~Chen, L.~Lin, and X.~Liang, ``Logicsolver: Towards interpretable math word problem solving with logical prompt-enhanced learning,'' in \emph{EMNLP}, 2022, pp. 1--13.

\bibitem{liu2023guiding}
J.~Liu, Z.~Huang, Z.~Ma, Q.~Liu, E.~Chen, T.~Su, and H.~Liu, ``Guiding mathematical reasoning via mastering commonsense formula knowledge,'' in \emph{Proceedings of the 29th ACM SIGKDD Conference on Knowledge Discovery and Data Mining}, 2023, pp. 1477--1488.

\bibitem{koncel2016mawps}
R.~Koncel-Kedziorski, S.~Roy, A.~Amini, N.~Kushman, and H.~Hajishirzi, ``Mawps: A math word problem repository,'' in \emph{Proceedings of the 2016 conference of the north american chapter of the association for computational linguistics: human language technologies}, 2016, pp. 1152--1157.

\bibitem{brown2020language}
T.~Brown, B.~Mann, N.~Ryder, M.~Subbiah, J.~D. Kaplan, P.~Dhariwal, A.~Neelakantan, P.~Shyam, G.~Sastry, A.~Askell \emph{et~al.}, ``Language models are few-shot learners,'' \emph{Advances in neural information processing systems}, vol.~33, pp. 1877--1901, 2020.

\bibitem{xie2019goal}
Z.~Xie and S.~Sun, ``A goal-driven tree-structured neural model for math word problems.'' in \emph{IJCAI}, 2019, pp. 5299--5305.

\bibitem{wang2018translating}
L.~Wang, Y.~Wang \emph{et~al.}, ``Translating a math word problem to a expression tree,'' in \emph{EMNLP}, 2018, pp. 1064--1069.

\bibitem{wang2022structure}
B.~Wang, J.~Ju, Y.~Fan \emph{et~al.}, ``Structure-unified m-tree coding solver for math word problem,'' in \emph{EMNLP}, 2022.

\bibitem{cao2021bottom}
Y.~Cao, F.~Hong, H.~Li, and P.~Luo, ``A bottom-up dag structure extraction model for math word problems,'' in \emph{AAAI}, vol.~35, no.~1, 2021, pp. 39--46.

\bibitem{wei2022chain}
J.~Wei, X.~Wang, D.~Schuurmans, M.~Bosma, F.~Xia, E.~Chi, Q.~V. Le, D.~Zhou \emph{et~al.}, ``Chain-of-thought prompting elicits reasoning in large language models,'' \emph{Advances in neural information processing systems}, vol.~35, pp. 24\,824--24\,837, 2022.

\bibitem{zhao2023survey}
W.~X. Zhao, K.~Zhou, J.~Li, T.~Tang, X.~Wang, Y.~Hou, Y.~Min, B.~Zhang, J.~Zhang, Z.~Dong \emph{et~al.}, ``A survey of large language models,'' \emph{arXiv preprint arXiv:2303.18223}, 2023.

\bibitem{wang2023document}
L.~Wang, C.~Lyu, T.~Ji, Z.~Zhang, D.~Yu, S.~Shi, and Z.~Tu, ``Document-level machine translation with large language models,'' in \emph{The 2023 Conference on Empirical Methods in Natural Language Processing}, 2023.

\bibitem{zhang2024vision}
J.~Zhang, J.~Huang, S.~Jin, and S.~Lu, ``Vision-language models for vision tasks: A survey,'' \emph{IEEE Transactions on Pattern Analysis and Machine Intelligence}, 2024.

\bibitem{zhu-etal-2023-solving}
\BIBentryALTinterwordspacing
X.~Zhu, J.~Wang, L.~Zhang, Y.~Zhang, Y.~Huang, R.~Gan, J.~Zhang, and Y.~Yang, ``Solving math word problems via cooperative reasoning induced language models,'' in \emph{Proceedings of the 61st Annual Meeting of the Association for Computational Linguistics (Volume 1: Long Papers)}, A.~Rogers, J.~Boyd-Graber, and N.~Okazaki, Eds.\hskip 1em plus 0.5em minus 0.4em\relax Toronto, Canada: Association for Computational Linguistics, Jul. 2023, pp. 4471--4485. [Online]. Available: \url{https://aclanthology.org/2023.acl-long.245}
\BIBentrySTDinterwordspacing

\bibitem{liang-etal-2023-gpt}
\BIBentryALTinterwordspacing
Z.~Liang, W.~Yu, T.~Rajpurohit, P.~Clark, X.~Zhang, and A.~Kalyan, ``Let {GPT} be a math tutor: Teaching math word problem solvers with customized exercise generation,'' in \emph{Proceedings of the 2023 Conference on Empirical Methods in Natural Language Processing}, H.~Bouamor, J.~Pino, and K.~Bali, Eds.\hskip 1em plus 0.5em minus 0.4em\relax Singapore: Association for Computational Linguistics, Dec. 2023, pp. 14\,384--14\,396. [Online]. Available: \url{https://aclanthology.org/2023.emnlp-main.889}
\BIBentrySTDinterwordspacing

\bibitem{lu2023survey}
P.~Lu, L.~Qiu, W.~Yu, S.~Welleck, and K.-W. Chang, ``A survey of deep learning for mathematical reasoning,'' in \emph{The 61st Annual Meeting Of The Association For Computational Linguistics}, 2023.

\bibitem{zador2023catalyzing}
A.~Zador, S.~Escola, B.~Richards, B.~{\"O}lveczky, Y.~Bengio, K.~Boahen, M.~Botvinick, D.~Chklovskii, A.~Churchland, C.~Clopath \emph{et~al.}, ``Catalyzing next-generation artificial intelligence through neuroai,'' \emph{Nature communications}, vol.~14, no.~1, p. 1597, 2023.

\bibitem{yao2024tree}
S.~Yao, D.~Yu, J.~Zhao \emph{et~al.}, ``Tree of thoughts: Deliberate problem solving with large language models,'' \emph{Advances in Neural Information Processing Systems}, vol.~36, 2024.

\bibitem{besta2024graph}
M.~Besta, N.~Blach, A.~Kubicek \emph{et~al.}, ``Graph of thoughts: Solving elaborate problems with large language models,'' in \emph{AAAI}, vol.~38, no.~16, 2024, pp. 17\,682--17\,690.

\bibitem{chen2023program}
W.~Chen, X.~Ma, X.~Wang, and W.~W. Cohen, ``Program of thoughts prompting: Disentangling computation from reasoning for numerical reasoning tasks,'' \emph{Transactions on Machine Learning Research}, 2023.

\bibitem{gao2023pal}
L.~Gao, A.~Madaan, S.~Zhou, U.~Alon, P.~Liu, Y.~Yang, J.~Callan, and G.~Neubig, ``Pal: Program-aided language models,'' in \emph{International Conference on Machine Learning}.\hskip 1em plus 0.5em minus 0.4em\relax PMLR, 2023, pp. 10\,764--10\,799.

\bibitem{mayer2012process}
R.~E. Mayer and M.~Hegarty, ``The process of understanding mathematical problems,'' in \emph{The nature of mathematical thinking}.\hskip 1em plus 0.5em minus 0.4em\relax Routledge, 2012, pp. 29--53.

\bibitem{phonapichat2014analysis}
P.~Phonapichat, S.~Wongwanich, and S.~Sujiva, ``An analysis of elementary school students’ difficulties in mathematical problem solving,'' \emph{Procedia-social and behavioral sciences}, vol. 116, pp. 3169--3174, 2014.

\bibitem{daroczy2015word}
G.~Daroczy, M.~Wolska, W.~D. Meurers, and H.-C. Nuerk, ``Word problems: A review of linguistic and numerical factors contributing to their difficulty,'' \emph{Frontiers in psychology}, vol.~6, p. 348, 2015.

\bibitem{pongsakdi2020makes}
N.~Pongsakdi, A.~Kajamies, K.~Veermans, K.~Lertola, M.~Vauras, and E.~Lehtinen, ``What makes mathematical word problem solving challenging? exploring the roles of word problem characteristics, text comprehension, and arithmetic skills,'' \emph{Zdm}, vol.~52, pp. 33--44, 2020.

\bibitem{anderson2014human}
J.~R. Anderson and G.~H. Bower, \emph{Human associative memory}.\hskip 1em plus 0.5em minus 0.4em\relax Psychology press, 2014.

\bibitem{raaijmakers1981search}
J.~G. Raaijmakers and R.~M. Shiffrin, ``Search of associative memory.'' \emph{Psychological review}, vol.~88, no.~2, p.~93, 1981.

\bibitem{johnson2002neural}
S.~C. Johnson, L.~C. Baxter, L.~S. Wilder, J.~G. Pipe, J.~E. Heiserman, and G.~P. Prigatano, ``Neural correlates of self-reflection,'' \emph{Brain}, vol. 125, no.~8, pp. 1808--1814, 2002.

\bibitem{grant2002self}
A.~M. Grant, J.~Franklin, and P.~Langford, ``The self-reflection and insight scale: A new measure of private self-consciousness,'' \emph{Social Behavior and Personality: an international journal}, vol.~30, no.~8, pp. 821--835, 2002.

\bibitem{astington201316}
J.~W. Astington and C.~Hughes, ``16 theory of mind: Self-reflection and social understanding,'' \emph{The Oxford handbook of developmental psychology, Vol. 2: Self and other}, vol.~2, p. 398, 2013.

\bibitem{hospedales2021meta}
T.~Hospedales, A.~Antoniou, P.~Micaelli, and A.~Storkey, ``Meta-learning in neural networks: A survey,'' \emph{IEEE Transactions on Pattern Analysis and Machine Intelligence}, vol.~44, no.~9, pp. 5149--5169, 2021.

\bibitem{guo2021context}
D.~Guo, H.~Wang, and M.~Wang, ``Context-aware graph inference with knowledge distillation for visual dialog,'' \emph{IEEE Transactions on Pattern Analysis and Machine Intelligence}, vol.~44, no.~10, pp. 6056--6073, 2021.

\bibitem{liu2022cognitive}
J.~Liu, Z.~Huang, X.~Lin, Q.~Liu, J.~Ma, and E.~Chen, ``A cognitive solver with autonomously knowledge learning for reasoning mathematical answers,'' in \emph{2022 IEEE International Conference on Data Mining (ICDM)}.\hskip 1em plus 0.5em minus 0.4em\relax IEEE, 2022, pp. 269--278.

\bibitem{kojima2022large}
T.~Kojima, S.~S. Gu, M.~Reid, Y.~Matsuo, and Y.~Iwasawa, ``Large language models are zero-shot reasoners,'' \emph{Advances in neural information processing systems}, vol.~35, pp. 22\,199--22\,213, 2022.

\bibitem{zhouleast}
D.~Zhou, N.~Sch{\"a}rli, L.~Hou, J.~Wei, N.~Scales, X.~Wang, D.~Schuurmans, C.~Cui, O.~Bousquet, Q.~V. Le \emph{et~al.}, ``Least-to-most prompting enables complex reasoning in large language models,'' in \emph{The Eleventh International Conference on Learning Representations}, 2023.

\bibitem{zhang2021teacher}
J.~Zhang, R.~K.-W. Lee, E.-P. Lim, W.~Qin, L.~Wang, J.~Shao, and Q.~Sun, ``Teacher-student networks with multiple decoders for solving math word problem,'' in \emph{Proceedings of the Twenty-Ninth International Conference on International Joint Conferences on Artificial Intelligence}, 2021, pp. 4011--4017.

\bibitem{wu2021math}
Q.~Wu, Q.~Zhang, Z.~Wei, and X.-J. Huang, ``Math word problem solving with explicit numerical values,'' in \emph{Proceedings of the 59th Annual Meeting of the Association for Computational Linguistics and the 11th International Joint Conference on Natural Language Processing (Volume 1: Long Papers)}, 2021, pp. 5859--5869.

\bibitem{zhang2022hgen}
Y.~Zhang, G.~Zhou, Z.~Xie, and J.~X. Huang, ``Hgen: Learning hierarchical heterogeneous graph encoding for math word problem solving,'' \emph{IEEE/ACM Transactions on Audio, Speech, and Language Processing}, vol.~30, pp. 816--828, 2022.

\bibitem{shen2020solving}
Y.~Shen and C.~Jin, ``Solving math word problems with multi-encoders and multi-decoders,'' in \emph{Proceedings of the 28th International Conference on Computational Linguistics}, 2020, pp. 2924--2934.

\bibitem{wu2021edge}
Q.~Wu, Q.~Zhang, and Z.~Wei, ``An edge-enhanced hierarchical graph-to-tree network for math word problem solving,'' in \emph{EMNLP 2021}, 2021, pp. 1473--1482.

\bibitem{roy2018mapping}
S.~Roy and D.~Roth, ``Mapping to declarative knowledge for word problem solving,'' \emph{Transactions of the Association for Computational Linguistics}, vol.~6, pp. 159--172, 2018.

\bibitem{devlin2019bert}
J.~Devlin, M.-W. Chang, K.~Lee, and K.~Toutanova, ``Bert: Pre-training of deep bidirectional transformers for language understanding,'' in \emph{Proceedings of the 2019 Conference of the North American Chapter of the Association for Computational Linguistics: Human Language Technologies, Volume 1 (Long and Short Papers)}, 2019, pp. 4171--4186.

\bibitem{Liu2019RoBERTaAR}
Y.~Liu, M.~Ott, N.~Goyal, J.~Du, M.~Joshi, D.~Chen, O.~Levy, M.~Lewis, L.~Zettlemoyer, and V.~Stoyanov, ``Roberta: A robustly optimized bert pretraining approach,'' \emph{arXiv preprint arXiv:1907.11692}, 2019.

\bibitem{petroni2019language}
F.~Petroni, T.~Rockt{\"a}schel, S.~Riedel, P.~Lewis, A.~Bakhtin, Y.~Wu, and A.~Miller, ``Language models as knowledge bases?'' in \emph{Proceedings of the 2019 Conference on Empirical Methods in Natural Language Processing and the 9th International Joint Conference on Natural Language Processing (EMNLP-IJCNLP)}, 2019, pp. 2463--2473.

\bibitem{kim2022improving}
H.~Kim, J.~Hwang, T.~Yoo, and Y.-G. Cheong, ``Improving a graph-to-tree model for solving math word problems,'' in \emph{2022 16th International Conference on Ubiquitous Information Management and Communication (IMCOM)}.\hskip 1em plus 0.5em minus 0.4em\relax IEEE, 2022, pp. 1--7.

\bibitem{yu2021improving}
W.~Yu, Y.~Wen, F.~Zheng, and N.~Xiao, ``Improving math word problems with pre-trained knowledge and hierarchical reasoning,'' in \emph{EMNLP}, 2021, pp. 3384--3394.

\bibitem{huang2021recall}
S.~Huang, J.~Wang, J.~Xu, D.~Cao, and M.~Yang, ``Recall and learn: A memory-augmented solver for math word problems,'' in \emph{Findings of the Association for Computational Linguistics: EMNLP 2021}, 2021, pp. 786--796.

\bibitem{liang2022mwp}
Z.~Liang, J.~Zhang, L.~Wang, W.~Qin, Y.~Lan, J.~Shao, and X.~Zhang, ``Mwp-bert: Numeracy-augmented pre-training for math word problem solving,'' in \emph{NAACL-HLT}, 2022, pp. 997--1009.

\bibitem{li2022seeking}
Z.~Li, W.~Zhang, C.~Yan, Q.~Zhou \emph{et~al.}, ``Seeking patterns, not just memorizing procedures: Contrastive learning for solving math word problems,'' in \emph{ACL}, 2022, pp. 2486--2496.

\bibitem{qin2023template}
J.~Qin, Z.~Yang, J.~Chen, X.~Liang, and L.~Lin, ``Template-based contrastive distillation pretraining for math word problem solving,'' \emph{IEEE Transactions on Neural Networks and Learning Systems}, 2023.

\bibitem{liu-etal-2019-tree}
Q.~Liu, W.~Guan, S.~Li, and D.~Kawahara, ``Tree-structured decoding for solving math word problems,'' in \emph{Proceedings of the 2019 Conference on Empirical Methods in Natural Language Processing and the 9th International Joint Conference on Natural Language Processing (EMNLP-IJCNLP)}, 2019, pp. 2370--2379.

\bibitem{hong2021learning}
Y.~Hong, Q.~Li, D.~Ciao, S.~Huang, and S.-C. Zhu, ``Learning by fixing: Solving math word problems with weak supervision,'' in \emph{Proceedings of the AAAI conference on artificial intelligence}, vol.~35, no.~6, 2021, pp. 4959--4967.

\bibitem{jie2022learning}
Z.~Jie, J.~Li, and W.~Lu, ``Learning to reason deductively: Math word problem solving as complex relation extraction,'' in \emph{ACL}, 2022, pp. 5944--5955.

\bibitem{zhang2022multi}
W.~Zhang, Y.~Shen, Y.~Ma, X.~Cheng, Z.~Tan, Q.~Nong, and W.~Lu, ``Multi-view reasoning: Consistent contrastive learning for math word problem,'' in \emph{Findings of the Association for Computational Linguistics: EMNLP 2022}, 2022, pp. 1103--1116.

\bibitem{wang2018mathdqn}
L.~Wang, D.~Zhang, L.~Gao, J.~Song, L.~Guo, and H.~T. Shen, ``Mathdqn: Solving arithmetic word problems via deep reinforcement learning,'' in \emph{Proceedings of the AAAI conference on artificial intelligence}, vol.~32, no.~1, 2018.

\bibitem{faldu2021towards}
K.~Faldu, A.~Sheth, P.~Kikani, M.~Gaur, and A.~Avasthi, ``Towards tractable mathematical reasoning: Challenges, strategies, and opportunities for solving math word problems,'' \emph{arXiv preprint arXiv:2111.05364}, 2021.

\bibitem{mnih2015human}
V.~Mnih, K.~Kavukcuoglu, D.~Silver, A.~A. Rusu, J.~Veness, M.~G. Bellemare, A.~Graves, M.~Riedmiller, A.~K. Fidjeland, G.~Ostrovski \emph{et~al.}, ``Human-level control through deep reinforcement learning,'' \emph{nature}, vol. 518, no. 7540, pp. 529--533, 2015.

\bibitem{mcclelland1995there}
J.~L. McClelland, B.~L. McNaughton, and R.~C. O'Reilly, ``Why there are complementary learning systems in the hippocampus and neocortex: insights from the successes and failures of connectionist models of learning and memory.'' \emph{Psychological review}, vol. 102, no.~3, p. 419, 1995.

\bibitem{kumaran2016learning}
D.~Kumaran, D.~Hassabis, and J.~L. McClelland, ``What learning systems do intelligent agents need? complementary learning systems theory updated,'' \emph{Trends in cognitive sciences}, vol.~20, no.~7, pp. 512--534, 2016.

\bibitem{o2014complementary}
R.~C. O’Reilly, R.~Bhattacharyya, M.~D. Howard, and N.~Ketz, ``Complementary learning systems,'' \emph{Cognitive science}, vol.~38, no.~6, pp. 1229--1248, 2014.

\bibitem{mikolov2013efficient}
T.~Mikolov, ``Efficient estimation of word representations in vector space,'' \emph{arXiv preprint arXiv:1301.3781}, 2013.

\bibitem{lin2023learning}
X.~Lin, Z.~Huang, H.~Zhao, E.~Chen, Q.~Liu, D.~Lian, X.~Li, and H.~Wang, ``Learning relation-enhanced hierarchical solver for math word problems,'' \emph{IEEE Transactions on Neural Networks and Learning Systems}, 2023.

\bibitem{velivckovic2018graph}
P.~Veli{\v{c}}kovi{\'c}, G.~Cucurull, A.~Casanova, A.~Romero, P.~Li{\`o}, and Y.~Bengio, ``Graph attention networks,'' in \emph{International Conference on Learning Representations}, 2018.

\bibitem{lai2011critical}
E.~R. Lai, ``Critical thinking: A literature review,'' \emph{Pearson's Research Reports}, vol.~6, no.~1, pp. 40--41, 2011.

\bibitem{moore2012critical}
B.~N. Moore and R.~Parker, \emph{Critical thinking}.\hskip 1em plus 0.5em minus 0.4em\relax McGraw-Hill, 2012.

\bibitem{paul1992critical}
R.~Paul and L.~Elder, ``Critical thinking: What, why, and how,'' \emph{New directions for community colleges}, vol.~77, no.~2, pp. 3--24, 1992.

\bibitem{hixon1993does}
J.~G. Hixon and W.~B. Swann, ``When does introspection bear fruit? self-reflection, self-insight, and interpersonal choices.'' \emph{Journal of personality and social psychology}, vol.~64, no.~1, p.~35, 1993.

\bibitem{marcovitch2008self}
S.~Marcovitch, S.~Jacques, J.~J. Boseovski, and P.~D. Zelazo, ``Self-reflection and the cognitive control of behavior: Implications for learning,'' \emph{Mind, Brain, and Education}, vol.~2, no.~3, pp. 136--141, 2008.

\bibitem{webb2023emergent}
T.~Webb, K.~J. Holyoak, and H.~Lu, ``Emergent analogical reasoning in large language models,'' \emph{Nature Human Behaviour}, vol.~7, no.~9, pp. 1526--1541, 2023.

\bibitem{sunstein1993analogical}
C.~R. Sunstein, ``On analogical reasoning,'' \emph{Harvard Law Review}, vol. 106, no.~3, pp. 741--791, 1993.

\bibitem{vosniadou1989similarity}
S.~Vosniadou and A.~Ortony, \emph{Similarity and analogical reasoning}.\hskip 1em plus 0.5em minus 0.4em\relax Cambridge University Press, 1989.

\bibitem{patel2021nlp}
A.~Patel, S.~Bhattamishra, and N.~Goyal, ``Are nlp models really able to solve simple math word problems?'' in \emph{Proceedings of the 2021 Conference of the North American Chapter of the Association for Computational Linguistics: Human Language Technologies}, 2021, pp. 2080--2094.

\bibitem{gaur2023reasoning}
V.~Gaur and N.~Saunshi, ``Reasoning in large language models through symbolic math word problems,'' in \emph{The 61st Annual Meeting Of The Association For Computational Linguistics}, 2023.

\bibitem{liang2023generalizing}
Z.~Liang, J.~Zhang, L.~Wang \emph{et~al.}, ``Generalizing math word problem solvers via solution diversification,'' in \emph{AAAI}, vol.~37, no.~11, 2023, pp. 13\,183--13\,191.

\bibitem{zhou2023learning}
Z.~Zhou, M.~Ning, Q.~Wang, J.~Yao, W.~Wang, X.~Huang, and K.~Huang, ``Learning by analogy: Diverse questions generation in math word problem,'' in \emph{61st Annual Meeting of the Association for Computational Linguistics, ACL 2023}.\hskip 1em plus 0.5em minus 0.4em\relax Association for Computational Linguistics (ACL), 2023, pp. 11\,091--11\,104.

\bibitem{steup2005epistemology}
M.~Steup and R.~Neta, ``Epistemology,'' 2005.

\bibitem{sloman2021cognitive}
S.~A. Sloman, R.~Patterson, and A.~K. Barbey, ``Cognitive neuroscience meets the community of knowledge,'' \emph{Frontiers in systems neuroscience}, vol.~15, p. 675127, 2021.

\bibitem{liu2023learning}
J.~Liu, Z.~Huang, C.~Zhai, and Q.~Liu, ``Learning by applying: A general framework for mathematical reasoning via enhancing explicit knowledge learning,'' in \emph{Proceedings of the AAAI Conference on Artificial Intelligence}, vol.~37, no.~4, 2023, pp. 4497--4506.

\bibitem{kahneman2011thinking}
D.~Kahneman, \emph{Thinking, fast and slow}.\hskip 1em plus 0.5em minus 0.4em\relax Macmillan, 2011.

\bibitem{evans2008dual}
J.~S.~B. Evans, ``Dual-processing accounts of reasoning, judgment, and social cognition,'' \emph{Annu. Rev. Psychol.}, vol.~59, pp. 255--278, 2008.

\bibitem{lieto2017dual}
A.~Lieto, D.~P. Radicioni, and V.~Rho, ``Dual peccs: a cognitive system for conceptual representation and categorization,'' \emph{Journal of Experimental \& Theoretical Artificial Intelligence}, vol.~29, no.~2, pp. 433--452, 2017.

\bibitem{atkinson1968human}
R.~C. Atkinson and R.~M. Shiffrin, ``Human memory: A proposed system and its control processes,'' in \emph{Psychology of learning and motivation}.\hskip 1em plus 0.5em minus 0.4em\relax Elsevier, 1968, vol.~2, pp. 89--195.

\bibitem{ccelikoz2019cognitive}
N.~{\c{C}}elik{\"o}z, Y.~Erisen, and M.~Sahin, ``Cognitive learning theories with emphasis on latent learning, gestalt and information processing theories.'' \emph{Online Submission}, vol.~9, no.~3, pp. 18--33, 2019.

\bibitem{simon1978information}
H.~A. Simon, ``Information-processing theory of human problem solving,'' \emph{Handbook of learning and cognitive processes}, vol.~5, pp. 271--295, 1978.

\bibitem{mowrer1960learning}
O.~Mowrer, ``Learning theory and behavior,'' 1960.

\bibitem{muhajirah2020basic}
M.~Muhajirah, ``Basic of learning theory: (behaviorism, cognitivism, constructivism, and humanism),'' \emph{International Journal of Asian Education}, vol.~1, no.~1, pp. 37--42, 2020.

\bibitem{kingma2013auto}
D.~P. Kingma and M.~Welling, ``Auto-encoding variational bayes,'' \emph{arXiv preprint arXiv:1312.6114}, 2013.

\bibitem{dai2023llm}
S.-C. Dai, A.~Xiong, and L.-W. Ku, ``Llm-in-the-loop: Leveraging large language model for thematic analysis,'' in \emph{The 2023 Conference on Empirical Methods in Natural Language Processing}.

\bibitem{liu2025socraticlm}
J.~Liu, Z.~Huang, T.~Xiao, J.~Sha, J.~Wu, Q.~Liu, S.~Wang, and E.~Chen, ``Socraticlm: Exploring socratic personalized teaching with large language models,'' \emph{Advances in Neural Information Processing Systems}, vol.~37, pp. 85\,693--85\,721, 2025.

\bibitem{mahowald2024dissociating}
K.~Mahowald, A.~A. Ivanova, I.~A. Blank, N.~Kanwisher, J.~B. Tenenbaum, and E.~Fedorenko, ``Dissociating language and thought in large language models,'' \emph{Trends in cognitive sciences}, 2024.

\bibitem{wang2010cognitive}
Y.~Wang and V.~Chiew, ``On the cognitive process of human problem solving,'' \emph{Cognitive systems research}, vol.~11, no.~1, pp. 81--92, 2010.

\bibitem{goldin1998representational}
G.~A. Goldin, ``Representational systems, learning, and problem solving in mathematics,'' \emph{The Journal of Mathematical Behavior}, vol.~17, no.~2, pp. 137--165, 1998.

\bibitem{xue2024decompose}
S.~Xue, Z.~Huang, J.~Liu, X.~Lin, Y.~Ning, B.~Jin, X.~Li, and Q.~Liu, ``Decompose, analyze and rethink: Solving intricate problems with human-like reasoning cycle,'' \emph{Advances in Neural Information Processing Systems}, vol.~37, pp. 357--385, 2024.

\bibitem{coda2023meta}
J.~Coda-Forno, M.~Binz, Z.~Akata, M.~Botvinick, J.~Wang, and E.~Schulz, ``Meta-in-context learning in large language models,'' \emph{Advances in Neural Information Processing Systems}, vol.~36, pp. 65\,189--65\,201, 2023.

\bibitem{wang2023large}
X.~Wang, W.~Zhu, M.~Saxon, M.~Steyvers, and W.~Y. Wang, ``Large language models are latent variable models: Explaining and finding good demonstrations for in-context learning,'' \emph{Advances in Neural Information Processing Systems}, vol.~36, pp. 15\,614--15\,638, 2023.

\bibitem{dong2022survey}
Q.~Dong, L.~Li, D.~Dai, C.~Zheng, Z.~Wu, B.~Chang, X.~Sun, J.~Xu, and Z.~Sui, ``A survey on in-context learning,'' \emph{arXiv preprint arXiv:2301.00234}, 2022.

\bibitem{liu2022makes}
J.~Liu, D.~Shen, Y.~Zhang, W.~B. Dolan, L.~Carin, and W.~Chen, ``What makes good in-context examples for gpt-3?'' in \emph{Proceedings of Deep Learning Inside Out (DeeLIO 2022): The 3rd Workshop on Knowledge Extraction and Integration for Deep Learning Architectures}, 2022, pp. 100--114.

\bibitem{luo2023dr}
M.~Luo, X.~Xu, Z.~Dai, P.~Pasupat, M.~Kazemi, C.~Baral, V.~Imbrasaite, and V.~Y. Zhao, ``Dr. icl: Demonstration-retrieved in-context learning,'' \emph{arXiv preprint arXiv:2305.14128}, 2023.

\bibitem{fu2022complexity}
Y.~Fu, H.~Peng, A.~Sabharwal, P.~Clark, and T.~Khot, ``Complexity-based prompting for multi-step reasoning,'' in \emph{The Eleventh International Conference on Learning Representations}, 2022.

\bibitem{robertson2009probabilistic}
S.~Robertson, H.~Zaragoza \emph{et~al.}, ``The probabilistic relevance framework: Bm25 and beyond,'' \emph{Foundations and Trends{\textregistered} in Information Retrieval}, vol.~3, no.~4, pp. 333--389, 2009.

\bibitem{raffel2020exploring}
C.~Raffel, N.~Shazeer, A.~Roberts, K.~Lee, S.~Narang, M.~Matena, Y.~Zhou, W.~Li, and P.~J. Liu, ``Exploring the limits of transfer learning with a unified text-to-text transformer,'' \emph{Journal of machine learning research}, vol.~21, no. 140, pp. 1--67, 2020.

\bibitem{chen2024bge}
J.~Chen, S.~Xiao, P.~Zhang \emph{et~al.}, ``Bge m3-embedding: Multi-lingual, multi-functionality, multi-granularity text embeddings through self-knowledge distillation,'' \emph{arXiv preprint arXiv:2402.03216}, 2024.

\bibitem{peng2024revisiting}
K.~Peng, L.~Ding, Y.~Yuan, X.~Liu, M.~Zhang, Y.~Ouyang, and D.~Tao, ``Revisiting demonstration selection strategies in in-context learning,'' in \emph{Proceedings of the 62nd Annual Meeting of the Association for Computational Linguistics (Volume 1: Long Papers)}, 2024, pp. 9090--9101.

\bibitem{van2024context}
M.-H. Van, X.~Wu \emph{et~al.}, ``In-context learning demonstration selection via influence analysis,'' \emph{arXiv preprint arXiv:2402.11750}, 2024.

\bibitem{nguyen2023context}
T.~Nguyen and E.~Wong, ``In-context example selection with influences,'' \emph{arXiv preprint arXiv:2302.11042}, 2023.

\bibitem{sorensen2022information}
T.~Sorensen, J.~Robinson, C.~Rytting, A.~Shaw, K.~Rogers, A.~Delorey, M.~Khalil, N.~Fulda, and D.~Wingate, ``An information-theoretic approach to prompt engineering without ground truth labels,'' in \emph{Proceedings of the 60th Annual Meeting of the Association for Computational Linguistics (Volume 1: Long Papers)}, 2022, pp. 819--862.

\bibitem{gonen2023demystifying}
H.~Gonen, S.~Iyer, T.~Blevins, N.~A. Smith, and L.~Zettlemoyer, ``Demystifying prompts in language models via perplexity estimation,'' in \emph{Findings of the Association for Computational Linguistics: EMNLP 2023}, 2023, pp. 10\,136--10\,148.

\bibitem{wu2023self}
Z.~Wu, Y.~Wang, J.~Ye, and L.~Kong, ``Self-adaptive in-context learning: An information compression perspective for in-context example selection and ordering,'' in \emph{Proceedings of the 61st Annual Meeting of the Association for Computational Linguistics (Volume 1: Long Papers)}, 2023, pp. 1423--1436.

\bibitem{qin2024context}
C.~Qin, A.~Zhang, C.~Chen, A.~Dagar, and W.~Ye, ``In-context learning with iterative demonstration selection,'' in \emph{Findings of the Association for Computational Linguistics: EMNLP 2024}, 2024, pp. 7441--7455.

\bibitem{liumakes}
J.~Liu, Z.~Huang, C.~Wang, X.~Huang, C.~Zhai, and E.~Chen, ``What makes in-context learning effective for mathematical reasoning,'' in \emph{Forty-second International Conference on Machine Learning}, 2025.

\bibitem{zhang2023retrieve}
P.~Zhang, S.~Xiao, Z.~Liu, Z.~Dou, and J.-Y. Nie, ``Retrieve anything to augment large language models,'' \emph{arXiv preprint arXiv:2310.07554}, 2023.

\bibitem{fan2024survey}
W.~Fan, Y.~Ding, L.~Ning, S.~Wang, H.~Li, D.~Yin, T.-S. Chua, and Q.~Li, ``A survey on rag meeting llms: Towards retrieval-augmented large language models,'' in \emph{Proceedings of the 30th ACM SIGKDD Conference on Knowledge Discovery and Data Mining}, 2024, pp. 6491--6501.

\bibitem{gao2023retrieval}
Y.~Gao, Y.~Xiong, X.~Gao, K.~Jia, J.~Pan, Y.~Bi, Y.~Dai, J.~Sun, H.~Wang, and H.~Wang, ``Retrieval-augmented generation for large language models: A survey,'' \emph{arXiv preprint arXiv:2312.10997}, vol.~2, 2023.

\bibitem{huang2024survey}
Y.~Huang and J.~Huang, ``A survey on retrieval-augmented text generation for large language models,'' \emph{arXiv preprint arXiv:2404.10981}, 2024.

\bibitem{ma2023query}
X.~Ma, Y.~Gong, P.~He, H.~Zhao, and N.~Duan, ``Query rewriting in retrieval-augmented large language models,'' in \emph{Proceedings of the 2023 Conference on Empirical Methods in Natural Language Processing}, 2023, pp. 5303--5315.

\bibitem{mao2024rafe}
S.~Mao, Y.~Jiang, B.~Chen, X.~Li, P.~Wang, X.~Wang, P.~Xie, F.~Huang, H.~Chen, and N.~Zhang, ``Rafe: Ranking feedback improves query rewriting for rag,'' in \emph{Findings of the Association for Computational Linguistics: EMNLP 2024}, 2024, pp. 884--901.

\bibitem{zhang2023repocoder}
F.~Zhang, B.~Chen, Y.~Zhang, J.~Keung, J.~Liu, D.~Zan, Y.~Mao, J.-G. Lou, and W.~Chen, ``Repocoder: Repository-level code completion through iterative retrieval and generation,'' in \emph{2023 Conference on Empirical Methods in Natural Language Processing (EMNLP 2023)}.\hskip 1em plus 0.5em minus 0.4em\relax Association for Computational Linguistics, 2023, pp. 2471--2484.

\bibitem{cheng2023lift}
X.~Cheng, D.~Luo, X.~Chen, L.~Liu, D.~Zhao, and R.~Yan, ``Lift yourself up: Retrieval-augmented text generation with self-memory,'' \emph{Advances in Neural Information Processing Systems}, vol.~36, pp. 43\,780--43\,799, 2023.

\bibitem{lewis2020retrieval}
P.~Lewis, E.~Perez, A.~Piktus, F.~Petroni, V.~Karpukhin, N.~Goyal, H.~K{\"u}ttler, M.~Lewis, W.-t. Yih, T.~Rockt{\"a}schel \emph{et~al.}, ``Retrieval-augmented generation for knowledge-intensive nlp tasks,'' \emph{Advances in neural information processing systems}, vol.~33, pp. 9459--9474, 2020.

\bibitem{goucritic}
Z.~Gou, Z.~Shao, Y.~Gong, Y.~Yang, N.~Duan, W.~Chen \emph{et~al.}, ``Critic: Large language models can self-correct with tool-interactive critiquing,'' in \emph{The Twelfth International Conference on Learning Representations}.

\bibitem{lan2024criticeval}
T.~Lan, W.~Zhang, C.~Xu, H.~Huang, D.~Lin, K.~Chen, and X.-L. Mao, ``Criticeval: Evaluating large-scale language model as critic,'' \emph{Advances in Neural Information Processing Systems}, vol.~37, pp. 66\,907--66\,960, 2024.

\bibitem{wang2022self}
X.~Wang, J.~Wei, D.~Schuurmans, Q.~V. Le, E.~H. Chi, S.~Narang, A.~Chowdhery, and D.~Zhou, ``Self-consistency improves chain of thought reasoning in language models,'' in \emph{The Eleventh International Conference on Learning Representations}, 2023.

\bibitem{weng2022large}
Y.~Weng, M.~Zhu, F.~Xia, B.~Li, S.~He, S.~Liu, B.~Sun, K.~Liu, and J.~Zhao, ``Large language models are better reasoners with self-verification,'' \emph{arXiv preprint arXiv:2212.09561}, 2022.

\bibitem{renze2024self}
M.~Renze and E.~Guven, ``Self-reflection in llm agents: Effects on problem-solving performance,'' \emph{arXiv preprint arXiv:2405.06682}, 2024.

\bibitem{gou2024critic}
\BIBentryALTinterwordspacing
Z.~Gou, Z.~Shao, Y.~Gong, yelong shen, Y.~Yang, N.~Duan, and W.~Chen, ``{CRITIC}: Large language models can self-correct with tool-interactive critiquing,'' in \emph{The Twelfth International Conference on Learning Representations}, 2024. [Online]. Available: \url{https://openreview.net/forum?id=Sx038qxjek}
\BIBentrySTDinterwordspacing

\bibitem{qintoolllm}
Y.~Qin, S.~Liang, Y.~Ye, K.~Zhu, L.~Yan, Y.~Lu, Y.~Lin, X.~Cong, X.~Tang, B.~Qian \emph{et~al.}, ``Toolllm: Facilitating large language models to master 16000+ real-world apis,'' in \emph{The Twelfth International Conference on Learning Representations}.

\bibitem{wangvoyager}
G.~Wang, Y.~Xie, Y.~Jiang, A.~Mandlekar, C.~Xiao, Y.~Zhu, L.~Fan, and A.~Anandkumar, ``Voyager: An open-ended embodied agent with large language models,'' \emph{Transactions on Machine Learning Research}.

\bibitem{ma2025advancing}
Z.~Ma, J.~Liu, X.~Luo, Z.~Huang, Q.~Zhu, and W.~Che, ``Advancing tool-augmented large language models via meta-verification and reflection learning,'' in \emph{Proceedings of the 31st ACM SIGKDD Conference on Knowledge Discovery and Data Mining V. 2}, 2025, pp. 2078--2089.

\bibitem{qian2023creator}
C.~Qian, C.~Han, Y.~R. Fung, Y.~Qin, Z.~Liu, and H.~Ji, ``Creator: Tool creation for disentangling abstract and concrete reasoning of large language models,'' in \emph{2023 Findings of the Association for Computational Linguistics: EMNLP 2023}.\hskip 1em plus 0.5em minus 0.4em\relax Association for Computational Linguistics (ACL), 2023, pp. 6922--6939.

\bibitem{yuancraft}
L.~Yuan, Y.~Chen, X.~Wang, Y.~Fung, H.~Peng, and H.~Ji, ``Craft: Customizing llms by creating and retrieving from specialized toolsets,'' in \emph{The Twelfth International Conference on Learning Representations}.

\bibitem{wang2024trove}
Z.~Z. Wang, G.~Neubig, and D.~Fried, ``Trove: inducing verifiable and efficient toolboxes for solving programmatic tasks,'' in \emph{Proceedings of the 41st International Conference on Machine Learning}, 2024, pp. 51\,177--51\,191.

\bibitem{ma2024automated}
Z.~Ma, Z.~Huang, J.~Liu, M.~Wang, H.~Zhao, and X.~Li, ``Automated creation of reusable and diverse toolsets for enhancing {LLM} reasoning,'' in \emph{The 39th Annual AAAI Conference on Artificial Intelligence}, 2024.

\bibitem{zhang-etal-2023-interpretable}
\BIBentryALTinterwordspacing
M.~Zhang, Z.~Wang, Z.~Yang, W.~Feng, and A.~Lan, ``Interpretable math word problem solution generation via step-by-step planning,'' in \emph{Proceedings of the 61st Annual Meeting of the Association for Computational Linguistics (Volume 1: Long Papers)}, A.~Rogers, J.~Boyd-Graber, and N.~Okazaki, Eds.\hskip 1em plus 0.5em minus 0.4em\relax Toronto, Canada: Association for Computational Linguistics, Jul. 2023, pp. 6858--6877. [Online]. Available: \url{https://aclanthology.org/2023.acl-long.379}
\BIBentrySTDinterwordspacing

\bibitem{sunsurvey}
\BIBentryALTinterwordspacing
J.~Sun, C.~Zheng, E.~Xie, Z.~Liu, R.~Chu, J.~Qiu, J.~Xu, M.~Ding, H.~Li, M.~Geng \emph{et~al.}, ``A survey of reasoning with foundation models,'' 2024. [Online]. Available: \url{https://arxiv.org/abs/2312.11562}
\BIBentrySTDinterwordspacing

\bibitem{hong2024advances}
R.~Hong, X.~Pang, and C.~Zhang, ``Advances in reasoning by prompting large language models: A survey,'' \emph{Cybernetics and Intelligence}, 2024.

\bibitem{yuemammoth}
X.~Yue, X.~Qu, G.~Zhang, Y.~Fu, W.~Huang, H.~Sun, Y.~Su, and W.~Chen, ``{MA}mmo{TH}: Building math generalist models through hybrid instruction tuning,'' in \emph{The Twelfth International Conference on Learning Representations}, 2024.

\bibitem{DBLP:conf/iclr/GouSGSYHDC24}
Z.~Gou, Z.~Shao, Y.~Gong, yelong shen, Y.~Yang, M.~Huang, N.~Duan, and W.~Chen, ``To{RA}: A tool-integrated reasoning agent for mathematical problem solving,'' in \emph{The Twelfth International Conference on Learning Representations}, 2024.

\bibitem{luo2023wizardmath}
H.~Luo, Q.~Sun, C.~Xu, P.~Zhao, J.~Lou, C.~Tao, X.~Geng, Q.~Lin, S.~Chen, and D.~Zhang, ``Wizardmath: Empowering mathematical reasoning for large language models via reinforced evol-instruct,'' \emph{arXiv preprint arXiv:2308.09583}, 2023.

\bibitem{xwin-lm}
\BIBentryALTinterwordspacing
X.-L. Team, ``Xwin-lm,'' 9 2023. [Online]. Available: \url{https://github.com/Xwin-LM/Xwin-LM}
\BIBentrySTDinterwordspacing

\bibitem{yu2023metamath}
L.~Yu, W.~Jiang, H.~Shi, J.~Yu, Z.~Liu, Y.~Zhang, J.~T. Kwok, Z.~Li, A.~Weller, and W.~Liu, ``Metamath: Bootstrap your own mathematical questions for large language models,'' \emph{arXiv preprint arXiv:2309.12284}, 2023.

\bibitem{cobbe2021trainingverifierssolvemath}
\BIBentryALTinterwordspacing
K.~Cobbe, V.~Kosaraju, M.~Bavarian, M.~Chen, H.~Jun, L.~Kaiser, M.~Plappert, J.~Tworek, J.~Hilton, R.~Nakano, C.~Hesse, and J.~Schulman, ``Training verifiers to solve math word problems,'' 2021. [Online]. Available: \url{https://arxiv.org/abs/2110.14168}
\BIBentrySTDinterwordspacing

\bibitem{huang2016well}
D.~Huang, S.~Shi, C.-Y. Lin, J.~Yin, and W.-Y. Ma, ``How well do computers solve math word problems? large-scale dataset construction and evaluation,'' in \emph{Proceedings of the 54th Annual Meeting of the Association for Computational Linguistics (Volume 1: Long Papers)}, 2016, pp. 887--896.

\bibitem{kushman2014learning}
N.~Kushman, Y.~Artzi, L.~Zettlemoyer, and R.~Barzilay, ``Learning to automatically solve algebra word problems,'' in \emph{Proceedings of the 52nd Annual Meeting of the Association for Computational Linguistics (Volume 1: Long Papers)}, 2014, pp. 271--281.

\bibitem{upadhyay2016learning}
S.~Upadhyay, M.-W. Chang, K.-W. Chang, and W.-t. Yih, ``Learning from explicit and implicit supervision jointly for algebra word problems,'' in \emph{Proceedings of the 2016 Conference on Empirical Methods in Natural Language Processing}, 2016, pp. 297--306.

\bibitem{hosseini2014learning}
M.~J. Hosseini, H.~Hajishirzi, O.~Etzioni, and N.~Kushman, ``Learning to solve arithmetic word problems with verb categorization,'' in \emph{Proceedings of the 2014 Conference on Empirical Methods in Natural Language Processing (EMNLP)}, 2014, pp. 523--533.

\bibitem{koncel2015parsing}
R.~Koncel-Kedziorski, H.~Hajishirzi, A.~Sabharwal, O.~Etzioni, and S.~D. Ang, ``Parsing algebraic word problems into equations,'' \emph{Transactions of the Association for Computational Linguistics}, vol.~3, pp. 585--597, 2015.

\bibitem{roy2015solving}
S.~Roy and D.~Roth, ``Solving general arithmetic word problems,'' in \emph{Proceedings of the 2015 Conference on Empirical Methods in Natural Language Processing}, 2015, pp. 1743--1752.

\bibitem{amini2019mathqa}
A.~Amini, S.~Gabriel, S.~Lin, R.~Koncel-Kedziorski, Y.~Choi, and H.~Hajishirzi, ``Mathqa: Towards interpretable math word problem solving with operation-based formalisms,'' in \emph{Proceedings of the 2019 Conference of the North American Chapter of the Association for Computational Linguistics: Human Language Technologies, Volume 1 (Long and Short Papers)}, 2019, pp. 2357--2367.

\bibitem{ling2017program}
W.~Ling, D.~Yogatama, C.~Dyer, and P.~Blunsom, ``Program induction by rationale generation: Learning to solve and explain algebraic word problems,'' in \emph{Proceedings of the 55th Annual Meeting of the Association for Computational Linguistics (Volume 1: Long Papers)}, 2017, pp. 158--167.

\bibitem{cobbe2021training}
K.~Cobbe, V.~Kosaraju, M.~Bavarian, M.~Chen, H.~Jun, L.~Kaiser, M.~Plappert, J.~Tworek, J.~Hilton, R.~Nakano \emph{et~al.}, ``Training verifiers to solve math word problems,'' \emph{arXiv preprint arXiv:2110.14168}, 2021.

\bibitem{xiao2024learning}
T.~Xiao, J.~Liu, Z.~Huang, J.~Wu, J.~Sha, S.~Wang, and E.~Chen, ``Learning to solve geometry problems via simulating human dual-reasoning process,'' in \emph{Proceedings of the Thirty-Third International Joint Conference on Artificial Intelligence}, 2024, pp. 6559--6568.

\bibitem{loveland2016automated}
D.~W. Loveland, \emph{Automated theorem proving: A logical basis}.\hskip 1em plus 0.5em minus 0.4em\relax Elsevier, 2016.

\bibitem{Seo2015SolvingGP}
M.~Seo, H.~Hajishirzi, A.~Farhadi, O.~Etzioni, and C.~Malcolm, ``Solving geometry problems: Combining text and diagram interpretation,'' in \emph{Proceedings of the 2015 conference on Empirical Methods in Natural Language Processing}, 2015, pp. 1466--1476.

\bibitem{Seo2014DiagramUI}
M.~J. Seo, H.~Hajishirzi, A.~Farhadi, and O.~Etzioni, ``Diagram understanding in geometry questions,'' in \emph{Proceedings of the AAAI Conference on Artificial Intelligence}, vol.~28, 2014.

\bibitem{Lu2021InterGPSIG}
P.~Lu, R.~Gong, S.~Jiang, L.~Qiu, S.~Huang, X.~Liang, and S.-c. Zhu, ``Inter-gps: Interpretable geometry problem solving with formal language and symbolic reasoning,'' in \emph{Proceedings of the 59th Annual Meeting of the Association for Computational Linguistics and the 11th International Joint Conference on Natural Language Processing (Volume 1: Long Papers)}, 2021, pp. 6774--6786.

\bibitem{Chen2021GeoQAAG}
J.~Chen, J.~Tang, J.~Qin, X.~Liang, L.~Liu, E.~Xing, and L.~Lin, ``Geoqa: A geometric question answering benchmark towards multimodal numerical reasoning,'' in \emph{Findings of the Association for Computational Linguistics: ACL-IJCNLP 2021}, 2021, pp. 513--523.

\bibitem{Cao2022AnAB}
J.~Cao and J.~Xiao, ``An augmented benchmark dataset for geometric question answering through dual parallel text encoding,'' in \emph{Proceedings of the 29th International Conference on Computational Linguistics}, 2022, pp. 1511--1520.

\bibitem{Ning2023ASC}
M.~Ning, Q.-F. Wang, K.~Huang, and X.~Huang, ``A symbolic characters aware model for solving geometry problems,'' in \emph{Proceedings of the 31st ACM International Conference on Multimedia}, 2023, pp. 7767--7775.

\bibitem{Zhang2022PlaneGD}
M.-L. Zhang, F.~Yin, Y.-H. Hao, and C.-L. Liu, ``Plane geometry diagram parsing,'' \emph{arXiv preprint arXiv:2205.09363}, 2022.

\bibitem{he2017mask}
K.~He, G.~Gkioxari, P.~Doll{\'a}r, and R.~Girshick, ``Mask r-cnn,'' in \emph{Proceedings of the IEEE International Conference on Computer Vision}, 2017, pp. 2961--2969.

\bibitem{ying2021embed}
\BIBentryALTinterwordspacing
H.~Ying, Z.~Huang, S.~Liu, T.~Shao, and K.~Zhou, ``Embedmask: Embedding coupling for instance segmentation,'' in \emph{International Joint Conference on Artificial Intelligence}, 2021, pp. 1266--1273. [Online]. Available: \url{https://doi.org/10.24963/ijcai.2021/175}
\BIBentrySTDinterwordspacing

\bibitem{xu2017scene}
D.~Xu, Y.~Zhu, C.~B. Choy, and L.~Fei-Fei, ``Scene graph generation by iterative message passing,'' in \emph{Proceedings of the IEEE/CVF Conference on Computer Vision and Pattern Recognition}, 2017, pp. 5410--5419.

\bibitem{Zhang2023AMN}
M.-L. Zhang, F.~Yin, and C.-L. Liu, ``A multi-modal neural geometric solver with textual clauses parsed from diagram,'' in \emph{International Joint Conference on Artificial Intelligence}, 2023.

\bibitem{trinh2024solving}
T.~H. Trinh, Y.~Wu, Q.~V. Le, H.~He, and T.~Luong, ``Solving olympiad geometry without human demonstrations,'' \emph{Nature}, vol. 625, no. 7995, pp. 476--482, 2024.

\bibitem{gao2023g}
J.~Gao, R.~Pi, J.~Zhang, J.~Ye, W.~Zhong, Y.~Wang, L.~Hong, J.~Han, H.~Xu, Z.~Li \emph{et~al.}, ``G-llava: Solving geometric problem with multi-modal large language model,'' \emph{arXiv preprint arXiv:2312.11370}, 2023.

\bibitem{zhang2024diagram}
Z.~Zhang, J.-K. Cheng, J.~Deng, L.~Tian, J.~Ma, Z.~Qin, X.~Zhang, N.~Zhu, and T.~Leng, ``Diagram formalization enhanced multi-modal geometry problem solver,'' \emph{arXiv preprint arXiv:2409.04214}, 2024.

\bibitem{schulz2002brainiac}
S.~Schulz, ``E--a brainiac theorem prover,'' \emph{Ai Communications}, vol.~15, no. 2-3, pp. 111--126, 2002.

\bibitem{kovacs2013first}
L.~Kov{\'a}cs and A.~Voronkov, ``First-order theorem proving and vampire,'' in \emph{International Conference on Computer Aided Verification}.\hskip 1em plus 0.5em minus 0.4em\relax Springer, 2013, pp. 1--35.

\bibitem{chou2000deductive}
S.-C. Chou, X.-S. Gao, and J.-Z. Zhang, ``A deductive database approach to automated geometry theorem proving and discovering,'' \emph{Journal of Automated Reasoning}, vol.~25, no.~3, pp. 219--246, 2000.

\bibitem{korovin2008iprover}
K.~Korovin, ``iprover--an instantiation-based theorem prover for first-order logic (system description),'' in \emph{International Joint Conference on Automated Reasoning}.\hskip 1em plus 0.5em minus 0.4em\relax Springer, 2008, pp. 292--298.

\bibitem{barbosa2022cvc5}
H.~Barbosa, C.~Barrett, M.~Brain, G.~Kremer, H.~Lachnitt, M.~Mann, A.~Mohamed, M.~Mohamed, A.~Niemetz, A.~N{\"o}tzli \emph{et~al.}, ``cvc5: A versatile and industrial-strength smt solver,'' in \emph{International Conference on Tools and Algorithms for the Construction and Analysis of Systems}.\hskip 1em plus 0.5em minus 0.4em\relax Springer, 2022, pp. 415--442.

\bibitem{irving2016deepmath}
G.~Irving, C.~Szegedy, A.~A. Alemi, N.~E{\'e}n, F.~Chollet, and J.~Urban, ``Deepmath-deep sequence models for premise selection,'' \emph{Advances in neural information processing systems}, vol.~29, 2016.

\bibitem{piotrowski2020stateful}
B.~Piotrowski and J.~Urban, ``Stateful premise selection by recurrent neural networks,'' \emph{arXiv preprint arXiv:2004.08212}, 2020.

\bibitem{prorokovic2021improving}
K.~Prorokovi{\'c}, M.~Wand, and J.~Schmidhuber, ``Improving stateful premise selection with transformers,'' in \emph{Intelligent Computer Mathematics: 14th International Conference, CICM 2021, Timisoara, Romania, July 26--31, 2021, Proceedings 14}.\hskip 1em plus 0.5em minus 0.4em\relax Springer, 2021, pp. 84--89.

\bibitem{wang2017premise}
M.~Wang, Y.~Tang, J.~Wang, and J.~Deng, ``Premise selection for theorem proving by deep graph embedding,'' \emph{Advances in neural information processing systems}, vol.~30, 2017.

\bibitem{paliwal2020graph}
A.~Paliwal, S.~Loos, M.~Rabe, K.~Bansal, and C.~Szegedy, ``Graph representations for higher-order logic and theorem proving,'' in \emph{Proceedings of the AAAI Conference on Artificial Intelligence}, vol.~34, no.~03, 2020, pp. 2967--2974.

\bibitem{lin2021contrastive}
Q.~Lin, J.~Liu, L.~Zhang, Y.~Pan, X.~Hu, F.~Xu, and H.~Zeng, ``Contrastive graph representations for logical formulas embedding,'' \emph{IEEE Transactions on Knowledge and Data Engineering}, vol.~35, no.~4, pp. 3563--3574, 2021.

\bibitem{ferreira2020natural}
D.~Ferreira and A.~Freitas, ``Natural language premise selection: Finding supporting statements for mathematical text,'' \emph{arXiv preprint arXiv:2004.14959}, 2020.

\bibitem{Ferreira2021STARC}
{Ferreira, Deborah and Freitas, Andr{\'e}}, ``Star: Cross-modal [sta]tement [r]epresentation for selecting relevant mathematical premises,'' in \emph{Conference of the European Chapter of the Association for Computational Linguistics}, 2021.

\bibitem{han2021proof}
J.~M. Han, J.~Rute, Y.~Wu, E.~W. Ayers, and S.~Polu, ``Proof artifact co-training for theorem proving with language models,'' \emph{arXiv preprint arXiv:2102.06203}, 2021.

\bibitem{huang2018gamepad}
D.~Huang, P.~Dhariwal, D.~Song, and I.~Sutskever, ``Gamepad: A learning environment for theorem proving,'' \emph{arXiv preprint arXiv:1806.00608}, 2018.

\bibitem{yang2019learning}
K.~Yang and J.~Deng, ``Learning to prove theorems via interacting with proof assistants,'' in \emph{International Conference on Machine Learning}.\hskip 1em plus 0.5em minus 0.4em\relax PMLR, 2019, pp. 6984--6994.

\bibitem{polu2020generative}
S.~Polu and I.~Sutskever, ``Generative language modeling for automated theorem proving,'' \emph{arXiv preprint arXiv:2009.03393}, 2020.

\bibitem{welleck2022naturalprover}
S.~Welleck, J.~Liu, X.~Lu, H.~Hajishirzi, and Y.~Choi, ``Naturalprover: Grounded mathematical proof generation with language models,'' \emph{Advances in Neural Information Processing Systems}, vol.~35, pp. 4913--4927, 2022.

\bibitem{jiang2022thor}
A.~Q. Jiang, W.~Li, S.~Tworkowski, K.~Czechowski, T.~Odrzyg{\'o}{\'z}d{\'z}, P.~Mi{\l}o{\'s}, Y.~Wu, and M.~Jamnik, ``Thor: Wielding hammers to integrate language models and automated theorem provers,'' \emph{Advances in Neural Information Processing Systems}, vol.~35, pp. 8360--8373, 2022.

\bibitem{zhang2023getting}
S.~D. Zhang, T.~Ringer, and E.~First, ``Getting more out of large language models for proofs,'' \emph{arXiv preprint arXiv:2305.04369}, 2023.

\bibitem{yousefzadeh2023large}
R.~Yousefzadeh and X.~Cao, ``Large language models' understanding of math: Source criticism and extrapolation,'' \emph{arXiv preprint arXiv:2311.07618}, 2023.

\bibitem{azerbayev2023llemma}
Z.~Azerbayev, H.~Schoelkopf, K.~Paster, M.~D. Santos, S.~McAleer, A.~Q. Jiang, J.~Deng, S.~Biderman, and S.~Welleck, ``Llemma: An open language model for mathematics,'' \emph{arXiv preprint arXiv:2310.10631}, 2023.

\bibitem{shao2402deepseekmath}
Z.~Shao, P.~Wang, Q.~Zhu, R.~Xu, J.~Song, X.~Bi, H.~Zhang, M.~Zhang, Y.~Li, Y.~Wu \emph{et~al.}, ``Deepseekmath: Pushing the limits of mathematical reasoning in open language models, 2024,'' \emph{URL https://arxiv. org/abs/2402}, vol. 3300, pp. 5--0.

\bibitem{frieder2024mathematical}
S.~Frieder, L.~Pinchetti, R.-R. Griffiths, T.~Salvatori, T.~Lukasiewicz, P.~Petersen, and J.~Berner, ``Mathematical capabilities of chatgpt,'' \emph{Advances in neural information processing systems}, vol.~36, 2024.

\bibitem{bansal2019holist}
K.~Bansal, S.~Loos, M.~Rabe, C.~Szegedy, and S.~Wilcox, ``Holist: An environment for machine learning of higher order logic theorem proving,'' in \emph{International Conference on Machine Learning}.\hskip 1em plus 0.5em minus 0.4em\relax PMLR, 2019, pp. 454--463.

\bibitem{rawson2019neurally}
M.~Rawson and G.~Reger, ``A neurally-guided, parallel theorem prover,'' in \emph{Frontiers of Combining Systems: 12th International Symposium, FroCoS 2019, London, UK, September 4-6, 2019, Proceedings 12}.\hskip 1em plus 0.5em minus 0.4em\relax Springer, 2019, pp. 40--56.

\bibitem{lample2205hypertree}
G.~Lample, M.-A. Lachaux, T.~Lavril, X.~Martinet, A.~Hayat, G.~Ebner, A.~Rodriguez, and T.~Lacroix, ``Hypertree proof search for neural theorem proving (2022),'' \emph{URL https://arxiv. org/abs/2205.11491}.

\bibitem{wang2023dt}
H.~Wang, Y.~Yuan, Z.~Liu, J.~Shen, Y.~Yin, J.~Xiong, E.~Xie, H.~Shi, Y.~Li, L.~Li \emph{et~al.}, ``Dt-solver: Automated theorem proving with dynamic-tree sampling guided by proof-level value function,'' in \emph{Proceedings of the 61st Annual Meeting of the Association for Computational Linguistics (Volume 1: Long Papers)}, 2023, pp. 12\,632--12\,646.

\bibitem{brandfonbrener2024vermcts}
D.~Brandfonbrener, S.~Henniger, S.~Raja, T.~Prasad, C.~R. Loughridge, F.~Cassano, S.~R. Hu, J.~Yang, W.~E. Byrd, R.~Zinkov \emph{et~al.}, ``Vermcts: Synthesizing multi-step programs using a verifier, a large language model, and tree search,'' in \emph{The 4th Workshop on Mathematical Reasoning and AI at NeurIPS'24}, 2024.

\bibitem{kusumoto2018automated}
M.~Kusumoto, K.~Yahata, and M.~Sakai, ``Automated theorem proving in intuitionistic propositional logic by deep reinforcement learning,'' \emph{arXiv preprint arXiv:1811.00796}, 2018.

\bibitem{fawzi2019learning}
A.~Fawzi, M.~Malinowski, H.~Fawzi, and O.~Fawzi, ``Learning dynamic polynomial proofs,'' \emph{Advances in Neural Information Processing Systems}, vol.~32, 2019.

\bibitem{wu2021tacticzero}
M.~Wu, M.~Norrish, C.~Walder, and A.~Dezfouli, ``Tacticzero: Learning to prove theorems from scratch with deep reinforcement learning,'' \emph{Advances in Neural Information Processing Systems}, vol.~34, pp. 9330--9342, 2021.

\bibitem{crouse2021deep}
M.~Crouse, I.~Abdelaziz, B.~Makni, S.~Whitehead, C.~Cornelio, P.~Kapanipathi, K.~Srinivas, V.~Thost, M.~Witbrock, and A.~Fokoue, ``A deep reinforcement learning approach to first-order logic theorem proving,'' in \emph{Proceedings of the AAAI Conference on Artificial Intelligence}, vol.~35, no.~7, 2021, pp. 6279--6287.

\bibitem{abdelaziz2022learning}
I.~Abdelaziz, M.~Crouse, B.~Makni, V.~Austel, C.~Cornelio, S.~Ikbal, P.~Kapanipathi, N.~Makondo, K.~Srinivas, M.~Witbrock \emph{et~al.}, ``Learning to guide a saturation-based theorem prover,'' \emph{IEEE Transactions on Pattern Analysis and Machine Intelligence}, vol.~45, no.~1, pp. 738--751, 2022.

\bibitem{fokoue2023ensemble}
A.~Fokoue, I.~Abdelaziz, M.~Crouse, S.~Ikbal, A.~Kishimoto, G.~Lima, N.~Makondo, and R.~Marinescu, ``An ensemble approach for automated theorem proving based on efficient name invariant graph neural representations,'' \emph{arXiv preprint arXiv:2305.08676}, 2023.

\bibitem{dabelow2025symbolic}
L.~Dabelow and M.~Ueda, ``Symbolic equation solving via reinforcement learning,'' \emph{Neurocomputing}, vol. 613, p. 128732, 2025.

\bibitem{wei2024proving}
C.~Wei, M.~Sun, and W.~Wang, ``Proving olympiad algebraic inequalities without human demonstrations,'' \emph{arXiv preprint arXiv:2406.14219}, 2024.

\bibitem{he2020machine}
Y.-H. He, K.-H. Lee, and T.~Oliver, ``Machine-learning number fields,'' \emph{arXiv preprint arXiv:2011.08958}, 2020.

\bibitem{lee2024exploring}
S.~Lee and S.~Kim, ``Exploring prime number classification: Achieving high recall rate and rapid convergence with sparse encoding,'' \emph{arXiv preprint arXiv:2402.03363}, 2024.

\bibitem{tong2024dart}
Y.~Tong, X.~Zhang, R.~Wang, R.~Wu, and J.~He, ``Dart-math: Difficulty-aware rejection tuning for mathematical problem-solving,'' \emph{arXiv preprint arXiv:2407.13690}, 2024.

\bibitem{hendrycks2021measuring}
D.~Hendrycks, C.~Burns, S.~Kadavath, A.~Arora, S.~Basart, E.~Tang, D.~Song, and J.~Steinhardt, ``Measuring mathematical problem solving with the math dataset,'' \emph{arXiv preprint arXiv:2103.03874}, 2021.

\end{thebibliography}

\newpage

\end{document}